%% file: main.tex
\RequirePackage{fix-cm}
\documentclass[twocolumn, final]{svjour3}          
\smartqed  

\makeatletter
\expandafter\let\csname opt@amsmath.sty\endcsname\relax
\AtBeginDocument{
  \mathindent=15pt 
  \@mathmargin\@centering} 
\makeatother

\usepackage{graphicx}
\UseRawInputEncoding

\usepackage[utf8]{inputenc}

\usepackage{times}
\usepackage{epsfig}
\usepackage{amsmath}
\usepackage{amssymb}
\usepackage{lipsum}
\usepackage{booktabs}
\newcommand{\R}{\mathbb{R}}
\usepackage{dsfont}
\usepackage{float}
\usepackage[caption = false]{subfig}
\usepackage{pifont}
\usepackage{multirow}
\usepackage[export]{adjustbox}
\usepackage[hyphens]{url}
\usepackage{hyperref}
\usepackage[hyphenbreaks]{breakurl}
\hypersetup{
    colorlinks=true,
    linkcolor=blue,
    filecolor=magenta,      
    urlcolor=cyan,
}
\usepackage{soul}

\newcommand{\cmark}{\ding{51}}
\journalname{Neural Computing and Applications}
\begin{document}




\title{Global-Local Attention for Emotion Recognition}




\author{Nhat Le$^{1, 2}$,
        Khanh Nguyen$^{1, 2}$,
        Anh Nguyen$^{3, *}$,
        Bac Le$^{1, 2}$
}

\authorrunning{Nhat Le \textit{et al.}}

\institute{$^{1}$ Faculty of Information Technology, University of Science, Ho Chi Minh City, Vietnam, \email{lhbac@fit.hcmus.edu.vn} \\
          $^{2}$ Vietnam National University, Ho Chi Minh City, Vietnam \\
            $^{3}$ Department of Computer Science, University of Liverpool, UK,
            \email{anh.nguyen@liverpool.ac.uk} \\
            $^{*}$ Corresponding author
}


\date{Received: date / Accepted: date}

\maketitle
\begin{abstract}
Human emotion recognition is an active research area in artificial intelligence and has made substantial progress over the past few years. Many recent works mainly focus on facial regions to infer human affection, while the surrounding context information is not effectively utilized. In this paper, we proposed a new deep network to effectively recognize human emotions using a novel global-local attention mechanism. Our network is designed to extract features from both facial and context regions independently, then learn them together using the attention module. In this way, both the facial and contextual information is used to infer human emotions, therefore enhancing the discrimination of the classifier. The intensive experiments show that our method surpasses the current state-of-the-art methods on recent emotion datasets by a fair margin. Qualitatively, our global-local attention module can extract more meaningful attention maps than previous methods. The source code and trained model of our network are available at \sloppy \burl{https://github.com/minhnhatvt/glamor-net}.


\end{abstract}
\input{1_intro}
\input{2_related_work}
\input{3_proposed_method}
\input{4_experiments}
\input{5_conclusion}

\section*{Declarations}

\hspace{3ex}\textbf{Conflict of interest}
The authors declare that they have no conflict of interest.

\textbf{Availability of data and material} The NCAER-S dataset can be downloaded at \url{https://bit.ly/NCAERS_dataset}.

\textbf{Code availability} The source code and trained model of our network are available at \url{https://github.com/minhnhatvt/glamor-net}.

\bibliographystyle{spmpsci}      
\bibliography{egbib}




\end{document}

%% file: 1_intro.tex
\section{Introduction}
Emotion recognition aims to classify input data into several expressions that convey universal emotions, such as \texttt{angry}, \texttt{disgust}, \texttt{fear}, \texttt{happy}, \texttt{neutral}, \texttt{sad}, and \texttt{surprise}. The input data can be one or more of different modalities such as visual information, audio, and text \cite{8014813} \cite{han2014speech} \cite{7374704}. Due to the availability of a large number of images and videos on the Internet, inferring human emotion from visual content, is considered to be one of the most popular tasks. Recently, automatic emotion recognition has gained a lot of attention in both academia and industry \cite{6940284}. It enables a wide range of novel applications in different domains, ranging from healthcare \cite{do2021multiple}, surveillance \cite{Author02} to robotics \cite{nguyen2020autonomous} and human-computer interaction \cite{Author03}. 
\begin{figure}[!t]
\captionsetup[subfigure]{labelformat=empty, farskip=1.5pt}
\centering
\setlength\tabcolsep{1.5pt} 
\begin{tabular}{cc}
\subfloat{\includegraphics[width=0.23\textwidth, height=3cm]{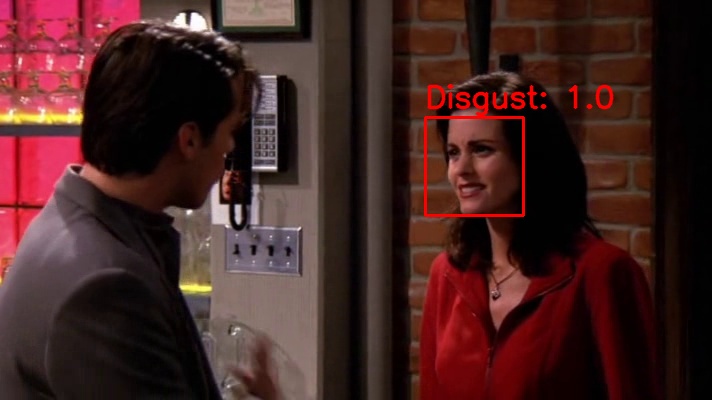}} &
\subfloat{\includegraphics[width=0.23\textwidth, height=3cm]{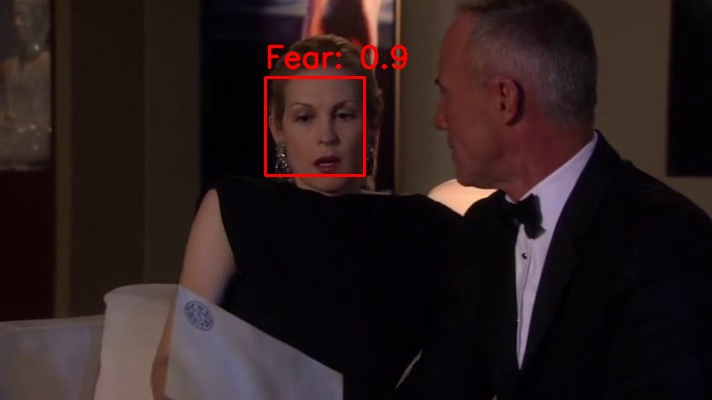}} \\
\subfloat{\includegraphics[width=0.23\textwidth, height=3cm]{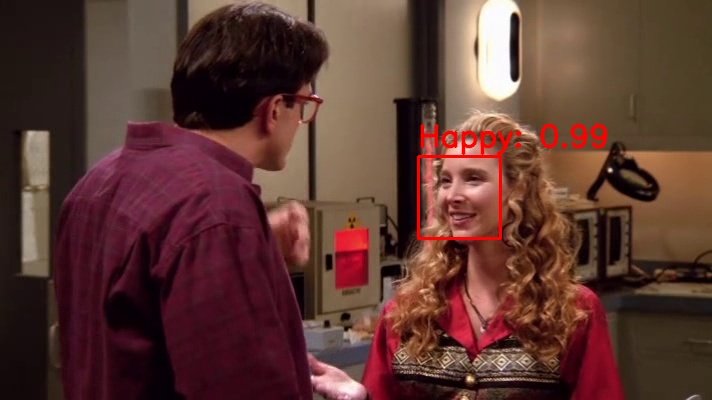}} &
\subfloat{\includegraphics[width=0.23\textwidth, height=3cm]{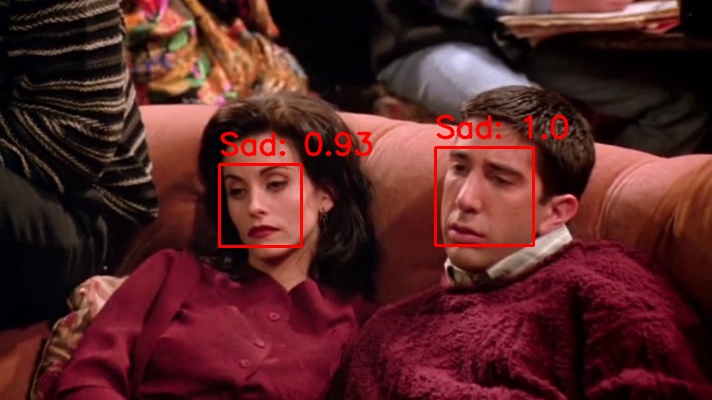}}
\end{tabular}
\caption{Examples of human emotion detection results from our method}
\label{fig:motivation}
\end{figure}

Traditional methods for emotion recognition combine handcrafted features (e.g, histogram of oriented gradients (HOG) \cite{hog}, local  binary  patterns) with classifiers such as SVM \cite{svm} or graphical models \cite{graphical_models}. With the popularity of deep learning techniques, especially Convolutional Neural Network (CNN) \cite{cnn}, together with the exists of many large-scale datasets, the meaningful features can be extracted using a deep network. However, the majority of previous methods \cite{onlyface1}\cite{onlyface2} \cite{onlyface3}\cite{FERpriormotion} only exploit features from human's face, and use this information to predict human emotions. These works assume that the facial region is the most informative representation of human emotion, therefore they ignore the surrounding context, which is shown to play an important role in the understanding of the perceived emotion, especially when the emotions on the face are expressed weakly or indistinguishable \cite{caer}. 

Recently, researchers have been focusing on incorporating background information such as people's pose, gaits, etc., into the model to improve the performance \cite{emoticon}\cite{binh2021graph}. In this work, we follow the same direction. However, unlike other works that learn the facial and context information independently \cite{emoticon}, we propose to jointly learn both facial and context information using our new Global-Local Attention mechanism. We hypothesize that the local information (i.e., facial region) and global information (i.e., context background) have a correlative relationship, and by simultaneously learning the attention using both of them, the accuracy of the network can be improved. This is based on the fact that the emotion of one person can be indicated by not only the face's emotion (i.e., local information) but also other context information such as the gesture, pose, or emotion/pose of a nearby person. Fig \ref{fig:motivation} shows some recognition results of our proposed method.

To verify the effectiveness of our approach, we benchmark on the CAER-S dataset \cite{caer}, a large-scale dataset for context-aware emotion recognition. We achieved $77.90\%$ top-1 accuracy on the test set, which is an improvement of $4.38\%$ over the recent state-of-the-art method \cite{caer}. Furthermore, with the integrated ResNet-18 {\cite{resnet}} as the backbone network, we obtained state-of-the-art performance on the CAER-S dataset with $\mathbf{89.88}\%$ classification accuracy. We also present a novel way to create a new static-image dataset from videos of the CAER dataset \cite{caer}. The experiments on this new dataset also confirm that our proposed method consistently achieves better performance than previous state-of-the-art approaches. 

In summary, our contributions are as follows:

\begin{itemize}
    \item We propose a new deep network, namely, \textbf{G}lobal-\textbf{L}ocal \textbf{A}ttention for E\textbf{mo}tion \textbf{R}ecognition \textbf{Net}work (GLAMOR-Net) that surpasses the state-of-the-art methods in the emotion recognition task.
    
    \item In GLAMOR-Net, we proposed the Global-Local Attention module, which successfully encodes both local features from facial regions and global features from surrounding background to improve the human emotion classification accuracy.
    
    \item We perform extensive experiments to validate the effectiveness of our proposed method and the contribution of each module on recent challenging datasets. 
\end{itemize}

The paper is organized as follow: We review the related work in Section \ref{sec:related_work}. We then describe our methodology in detail in Section \ref{sec:approach}. In Section \ref{sec:experiments}, we present extensive experimental results on challenging datasets and analyze the contribution of each module in GLAMOR-Net. Finally, we conclude the paper and discuss future work in Section \ref{sec_conclusion}.

%% file: 2_related_work.tex
\section{Related Work}
\label{sec:related_work}
\subsection{Human Emotion}
In the late twentieth century, Ekman and Friesen discovered six basic universal emotions including anger, disgust, fear, happiness, sadness, and surprise \cite{6emotions}. Several years later, contempt was added and considered as one of the basic emotions \cite{contempt}. However, our affective displays in reality are much more complicated and subtle compared to the simplicity of these universal emotions. To represent the complexity of the emotional spectrum, many approaches were proposed such as the Facial Action Coding System \cite{facs}, where all facial actions are described in terms of Action Units (AUs); or dimensional models \cite{dim_model}, where affection is quantified by values chosen over continuous emotional scales like valence and arousal. Nevertheless, those models which use discrete affections are the most popular in automatic emotion recognition task because they are easier to interpret and more intuitive to human.

\subsection{Emotion Recognition}
In automatic human emotion recognition, many approaches mainly focus on analyzing facial expression. Thus, a standard emotion recognition system usually consists of three main stages: face detection, feature extraction and expression classification \cite{onlyface1}\cite{onlyface2}\cite{onlyface3}\cite{FERpriormotion}). Traditional methods relied on handcrafted features (LBP\cite{LBP}, HOG\cite{hog}) to extract meaningful features from input images, and classifiers (such as SVM or random forest) to classify human emotions based on extracted features. With the rise of deep learning, CNN-based methods have made significant progress in the task of emotion recognition \cite{deepfer_survey}. Apart from using input image, other works focus on categorizing emotions by utilizing extra information such as speech \cite{speech_emotion1}\cite{speech_emotion2}, human pose \cite{body_emotion1}, body movements and gaits \cite{gait_emotion1}\cite{gait_emotion2}. However, these works have relied on the information coming from a single modality, hence they have limited ability to fully exploit all usable information of human emotions.

\begin{figure*}[t]
    \centering
    \includegraphics[width=0.98\textwidth]{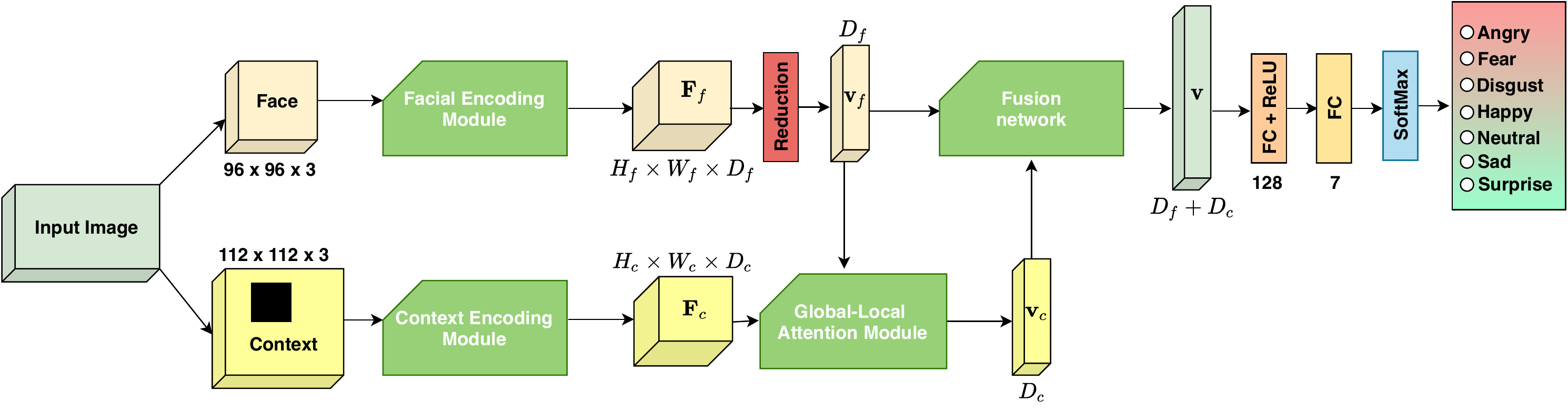}
    \vspace{2ex}
    \caption{The architecture of our proposed network. The whole process includes three steps. First, we extract the facial information (local) and context information (global) using two Encoding Modules. Second, we feed the extracted face and context features into the Global-Local Attention (GLA) module to perform attention inference on the global context. Lastly, we fuse both features from the facial region and output features from GLA into a neural network to make final emotion classification}
    \label{fig:full}
\end{figure*}

To overcome this limitation, many researches have investigated the use of multiple modalities. Primarily, these works tried to fuse multiple channels of information from each modality to predict emotion. Castellano \emph{et al.} \cite{multimodal1}  used extracted features from three different modalities (facial expressions, body gestures and speech expressions), and then fused those modalities at two different levels (i.e. feature level and decision level). Their results showed that the fusion performed at the feature level provided better results than the one performed at the decision level. Sikka \emph{et al.} \cite{multimodal2} extracted different visual features such as SIFT-Bag of Words \cite{sift_bow}, LPQ-TOP \cite{lpq_top}, HOG \cite{hog_features}, PHOG \cite{phog}, and GIST \cite{gist} and fuse them with audio features by building a kernel from each set of features, then combine them using a SVM classifier. Likewise, the authors in \cite{multimodal3} used the same multi-modality approach but using deep learning techniques. In \cite{emoticon}, three interpretations of context information are fused together by a deep neural network to classify human emotions in an end-to-end manner.

Recently, many works have focused on exploring context-aware information for emotion recognition. Kosti \emph{et al.} \cite{context_emotic} and Lee \emph{et al.} \cite{caer} proposed two architectures based on deep neural networks for learning context information. Both of them have two separate branches for extracting different kinds of information. One branch focuses on human features (i.e. face for \cite{caer} and body for \cite{context_emotic}) and the other concentrates on surrounding context. When considering multiple modalities, which have a large amount of information, deep learning-based methods like \cite{emoticon}\cite{caer}\cite{context_emotic}\cite{do2018affordancenet} are more suitable and effective than traditional approaches. These multi-modal approaches often yield better classification performance than uni-modal methods.

\subsection{Attention Model}
Attention was first introduced in machine translation \cite{machinetranslation}, allowing the translation model to search for words in the input sentence that are more relevant to the prediction words. Since then, attention models have become an important concept and an essential component of neural network architectures. It has made significant impacts in many application domains, including natural language processing \cite{NLPattention}, computer vision \cite{CVattention}, graph \cite{Graphattention} and speech processing \cite{Speechattention}.

In emotion recognition, attention models were mainly used to discover the attentive areas of the face that need to be focused on \cite{FERpriormotion}. Recently, the work that forced the model to pay attention to the most discriminative regions of the background using attention was proposed in CAER-Net-S \cite{caer}. However, previous work only used the background encoding to learn the context saliency map and did not take advantage of the facial representation to assist the process. Therefore, we propose the Global-Local Attention mechanism, which takes both facial and context encoding as inputs, to utilize facial information more efficiently to guide the context saliency map learning procedure.

%% file: 3_proposed_method.tex
\section{Methodology}
\label{sec:approach}
\subsection{Overview}
In this work, we assume that emotions can be recognized by understanding the context components of the scene together with the facial expression. Our method aims to do emotion recognition in the wild by incorporating both facial information of the person's face and contextual information surrounding that person. Our model consists of three components: Encoding Module, Global-Local Attention (GLA) Module, and Fusion Module. Our main contribution is the novel GLA module, which utilizes facial features as the local information to attend better to salient locations in the global context. Fig. \ref{fig:full} shows an overview of our method.  

\begin{figure*}
    \centering
    \includegraphics[width=0.62\textwidth]{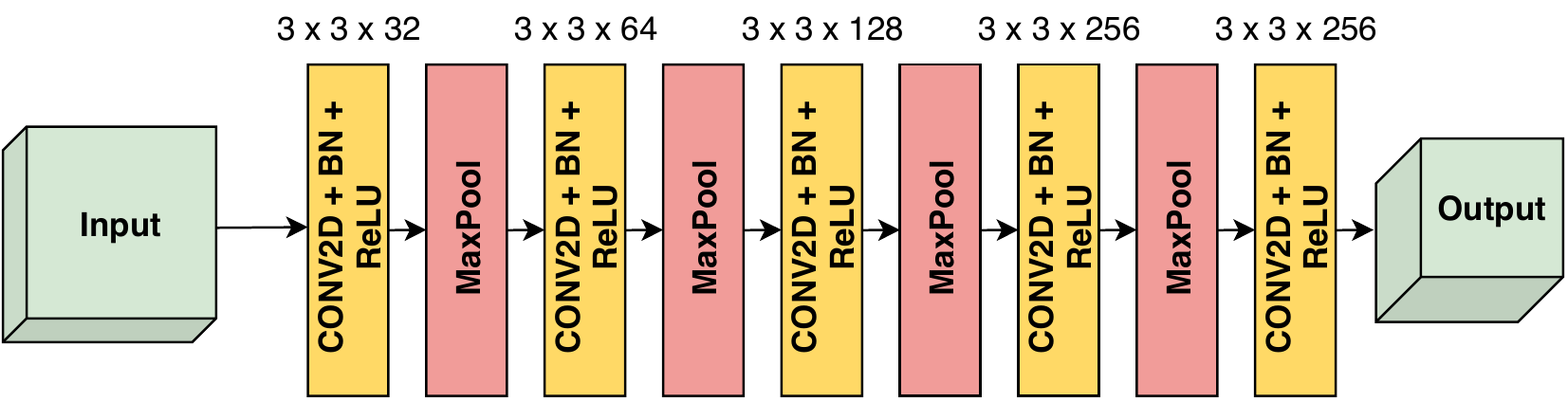}
    \vspace{2ex}
    \caption{Our proposed encoder network as the feature extractor for both face and context branches. The network contains five convolutional layers with ReLU non-linearity, each convolution is followed by a max pooling layer except the last one to reduce the spatial dimensions of the input}
    \label{fig:encoder}
\end{figure*}

\subsection{Network Architecture}
\subsubsection{Encoding Module}
\label{section:encoding_module}
To detect human emotion, many works first process the image by cropping out the human faces from the scene, and then feed them into a convolutional network to extract facially-expressive features \cite{onlyface1}\cite{onlyface2}\cite{onlyface3}\cite{FERpriormotion}. We generally follow this approach in our Encoding Module. In particular, our Encoding Module comprises the Facial Encoding Module to learn the face features, and the Context Encoding Module to learn the context features. 

\textbf{Facial Encoding Module} This module aims to learn meaningful features from the facial region of the input image. The facial embedding information can be denoted as $\mathbf{F}_f$:
\begin{equation}
    \mathbf{F}_f = \mathfrak{C}(\mathbf{I}_f; \theta_f)
\end{equation}
where $\mathfrak{C}$ is the convolutional operation parameterized by $\theta_f$, and $\mathbf{I}_f$ is the input facial region. In practice, we use a sub-network (Fig. \ref{fig:encoder}) as the feature extractor for the Facial Encoding Module. 

The proposed sub-network has five convolutional layers. Particularly, each convolutional layer has a kernel set of $3\times3$ filters with strides of $1\times1$ followed by a Batch Normalization layer and a ReLU activation function. The number of filters starts with 32 in the first layer, increasing by a factor of 2 at each subsequent layer except the last one. Our network ends up with 256 output channels. We also use the padding technique before each convolutional layer to keep the output spatial dimensions the same as the input. The output of each convolutional layer is pooled using a max-pooling layer with strides of $2\times2$. The encoding module outputs a 256-channel volume feature map, which is the embedded representation with respect to the input image.

\textbf{Context Encoding Module} This module is used to exploit background knowledge to support the emotion predicting process. Similar to the Facial Encoding Module, we follow the same procedure to extract context information contained in the scene with a different set of parameters:
\begin{equation}
    \mathbf{F}_c = \mathfrak{C}(\mathbf{I}_c; \theta_c)
\end{equation}
where $\mathfrak{C}$ is the convolutional operation parameterized by $\theta_c$, and $\mathbf{I}_c$ is the input context. Similar to the Facial Encoding Module, we use the sub-network (Fig. \ref{fig:encoder}) to extract deep features from the background context region in the Context Encoding Module. 

After getting these two feature maps, we feed them into the Global-Local Attention Module to calculate the attention scores for regions in the context. However, if we extract the context information in the raw image where the faces apparently exist, the network will also encode the facial information. This problem can make the attention module produce trivial outputs because the network may only focus on the facial region, and omitting the context information in other parts of the image. To address this problem, we first detect the face and then hide it in the raw input by setting all the values in the facial region to zero.

\subsubsection{Global-Local Attention Module}
\label{section:attention}

\begin{figure*}[]
    \centering
 \includegraphics[width=0.95\textwidth]{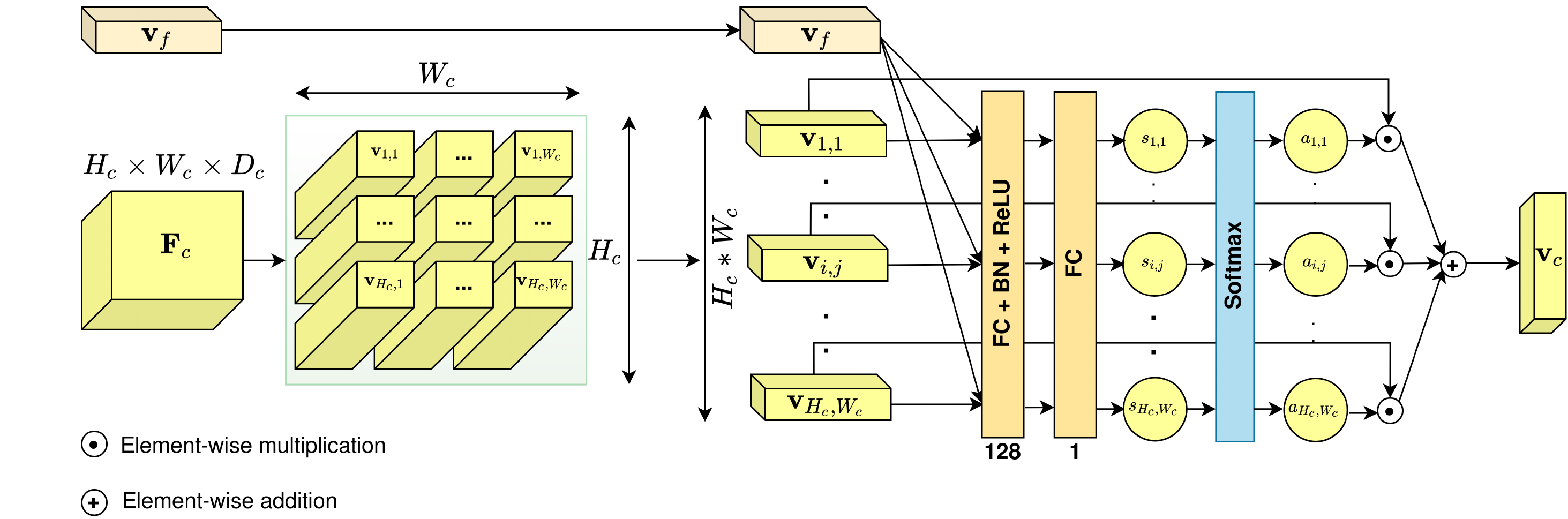}
    \vspace{2ex}
         \caption{The proposed Global-Local Attention module takes the extracted face feature vector and the context feature map as the input to perform context attention inference. Each vector $\mathbf{v}_{i,j}$ in the context feature map $\mathbf{F}_c$ is concatenated with the face vector $\mathbf{v}_f$ and then fed into a sub-network to compute the attention weight for the $(i,j)$ position. The final output vector is a linear combination of all regions in the context weighted by the corresponding attention weight. For efficiency, our attention inference network contains a 128-unit Fully Connected layer with the ReLU activation function and a Softmax layer. Weights are shared across all the context regions}
         \label{fig:attention_module}
\end{figure*}

\begin{figure*}
    \centering
    \includegraphics[width=0.55\textwidth]{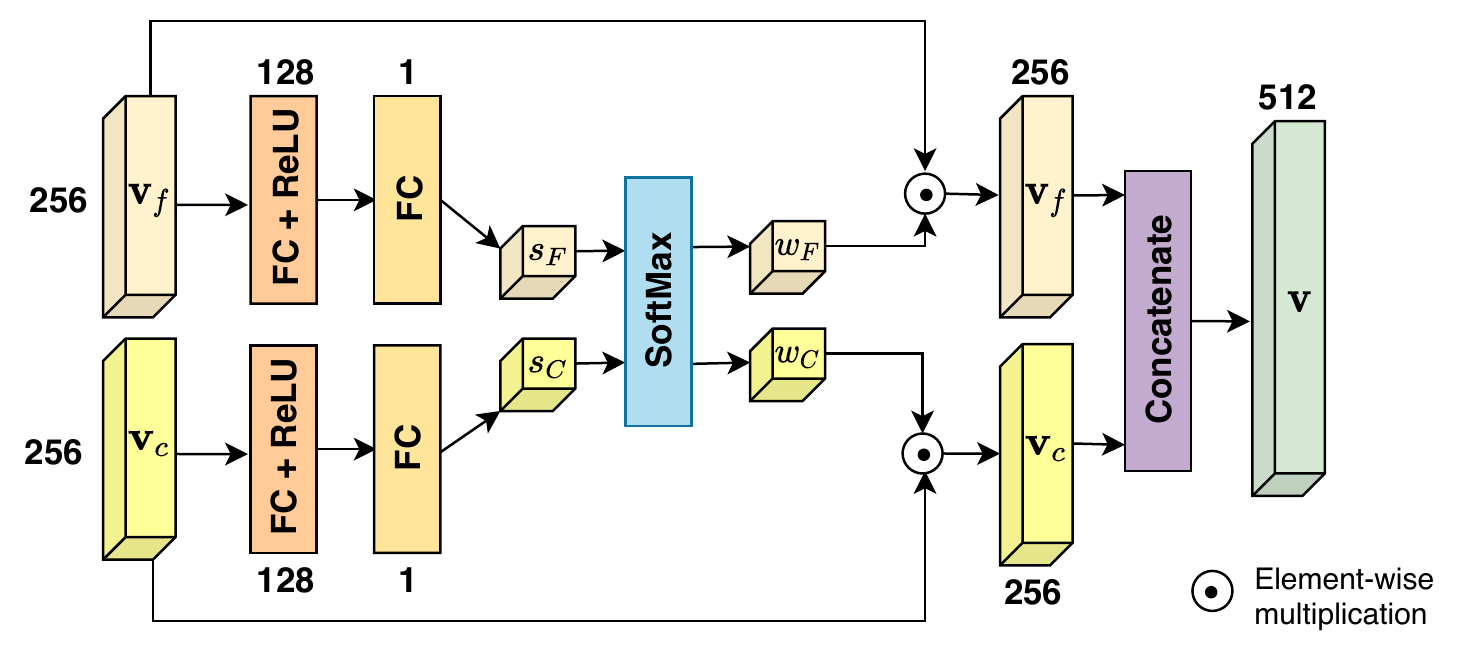}
    \vspace{2ex}
     \caption{The Fusion Module consists of two separate sub-networks, each network computes the fusion weights for face branch and context branch. The input vector of each branch is then scaled by its corresponding weight and combined together into the final representation vector $\mathbf{v}$. We use this vector $\mathbf{v}$ to estimate the emotion category by feeding it into another sub-network (see Fig. \ref{fig:full})}
    \label{fig:fusion}
\end{figure*}

Inspired by the attention mechanism \cite{Speechattention} \cite{nguyen2019v2cnet}, 
to model the associative relationship of the local information (i.e., the facial region in our work) and global information (i.e., the surrounding context background), we propose the Global-Local Attention Module to guide the network focus on meaningful regions (Fig. \ref{fig:attention_module}). Specifically, our attention mechanism models the hidden correlation between the face and different regions in the context by capturing their similarity using deep learning techniques. Our attention module takes the extracted face feature map $\mathbf{F}_f$ and the context feature map $\mathbf{F}_c$ from the two encoding modules as input, and then outputs a normalized saliency map that has the same spatial dimension as $\mathbf{F}_c$.

In practice, we first reduce the facial feature map $\mathbf{F}_f$ into vector representation using the Global Pooling operator, denoted as $\mathbf{v}_f$. Note that the context feature map $\mathbf{F}_c$ is a 3D tensor, $\mathbf{F}_c \in \R^{H_c \times W_c \times D_c}$, where $H_c$, $W_c$, and $D_c$ are the height, width, and channel dimension respectively. We derive the context feature map $\mathbf{F}_c$ as a set of $W_c * H_c$ vectors with $D_c$ dimensions, each vector in each cell $(i,j)$ represents the embedded features at that location, which can be projected back to the corresponding patch in the input image:
\begin{equation}
    \mathbf{F}_c =\{ \mathbf{v}_{i,j} \in \R^{D_c} |1 \leq i \leq H_c, 1 \leq j \leq W_c \}
\end{equation}
At each location $(i,j)$ in the context feature map, we have $\mathbf{F}_c^{(i,j)} = \mathbf{v}_{i,j}$, where $\mathbf{v}_{i,j} \in \R^{D_c}$ and $1 \leq i \leq H_c$, $1 \leq j \leq W_c$. 

We concatenate $[\mathbf{v}_f; \mathbf{v}_{i,j}]$ into a big vector $\bar{\mathbf{v}}_{i,j}$, which contains both information about the face and some small regions of the scene. We then employ a feed-forward neural network to compute the score corresponding to that region by feeding $\bar{\mathbf{v}}_{i,j}$ into the network. After repeating the same process for all regions, each region $(i,j)$ will output a raw score value $s_{i,j}$, we spatially apply the Softmax function to produce the attention map:
\begin{equation}
    a_{i,j} = \frac{\exp(s_{i,j})}{\Sigma_a\Sigma_b \exp(s_{a,b})}
\end{equation}
To obtain the final context representation vector, we squish the feature maps by taking the average over all the regions weighted by $a_{i,j}$ as follow:
\begin{equation}
    \mathbf{v}_c = \Sigma_i\Sigma_j(a_{i,j} \odot \mathbf{v}_{i,j})
\end{equation}
where $\mathbf{v}_c \in \R^{D_c}$ is the final single vector encoding the context information, and $\odot$ is the scalar multiplication operation. Additionally, $\mathbf{v}_c$ mainly contains information from regions that have high attention, while other unimportant parts of the context are mostly ignored. With this design, our attention module can guide the network focus on important areas based on both facial information and context information of the image. Note that, in practice, we only need to extract context information once and then using different encoded face representations to make the system look at different regions with respect to that person. 

\subsubsection{Fusion Module}
The Fusion Module is used to incorporate the facial and context information more effectively when predicting human emotions. The Fusion Module takes $\mathbf{v}_f$ and $\mathbf{v}_c$ as the input, then the face score and context score are computed independently by two neural networks:
\begin{align}
    s_f = \mathcal{F}(\mathbf{v}_f; \phi_f) && s_c = \mathcal{F}(\mathbf{v}_c; \phi_c)
\end{align}
where ${\phi_f}$ and $\phi_c$ are the network parameters of the face branch and context branch, respectively. Next, we normalize those scores by the Softmax function to produce weights for each face and context branch so that these weights sum up to 1. 
\begin{align}
    w_f = \frac{\exp(s_f)}{\exp(s_f)+\exp(s_c)} &&  w_c = \frac{\exp(s_c)}{\exp(s_f)+\exp(s_c)} 
\end{align}

\begin{figure*}
    \centering
    \setlength\tabcolsep{8pt} 
    \begin{tabular}{cc}
    \subfloat[CAER]{\includegraphics[width=0.35\textwidth]{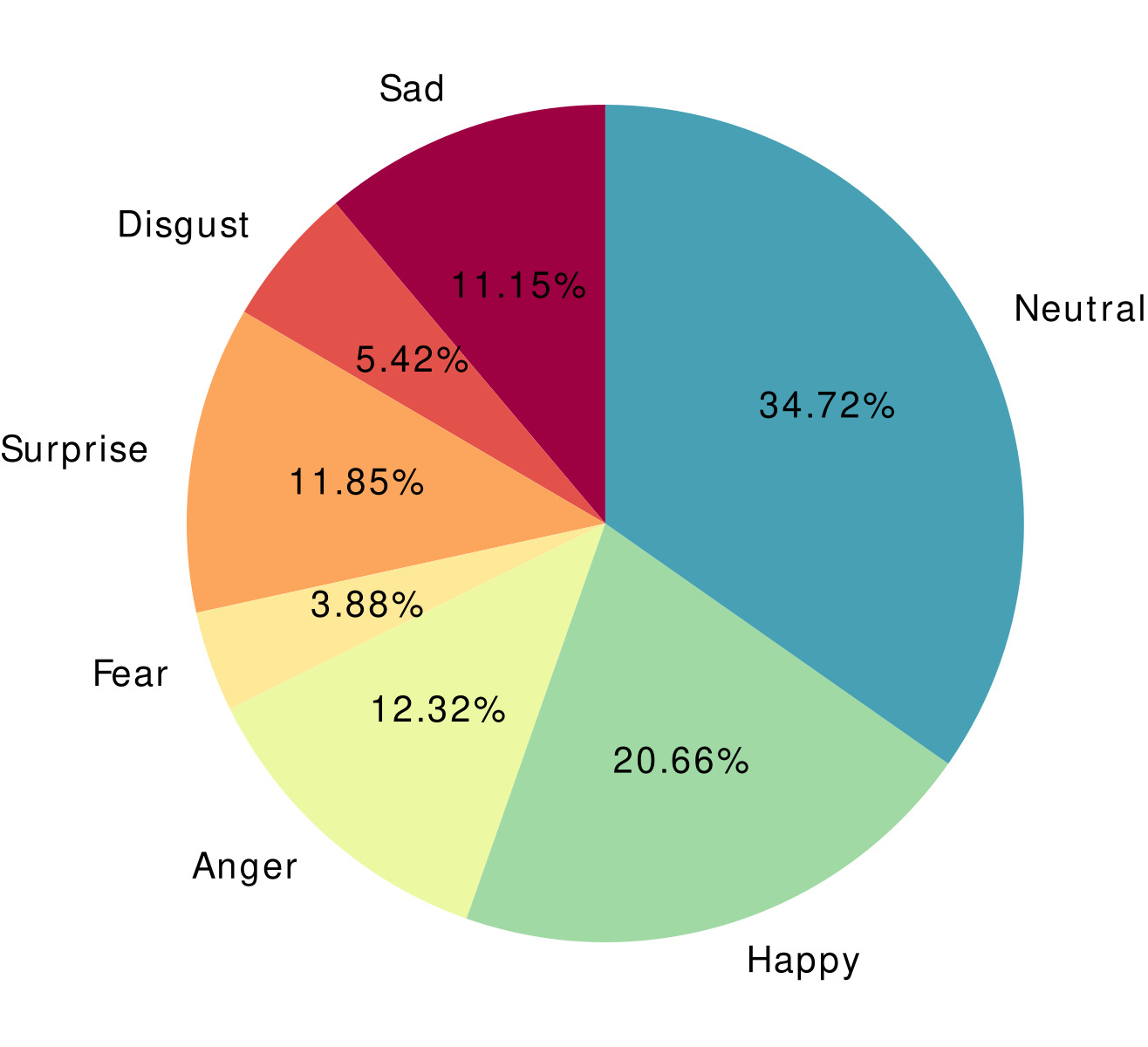}} &
    \subfloat[NCAER-S]{\includegraphics[width=0.35\textwidth]{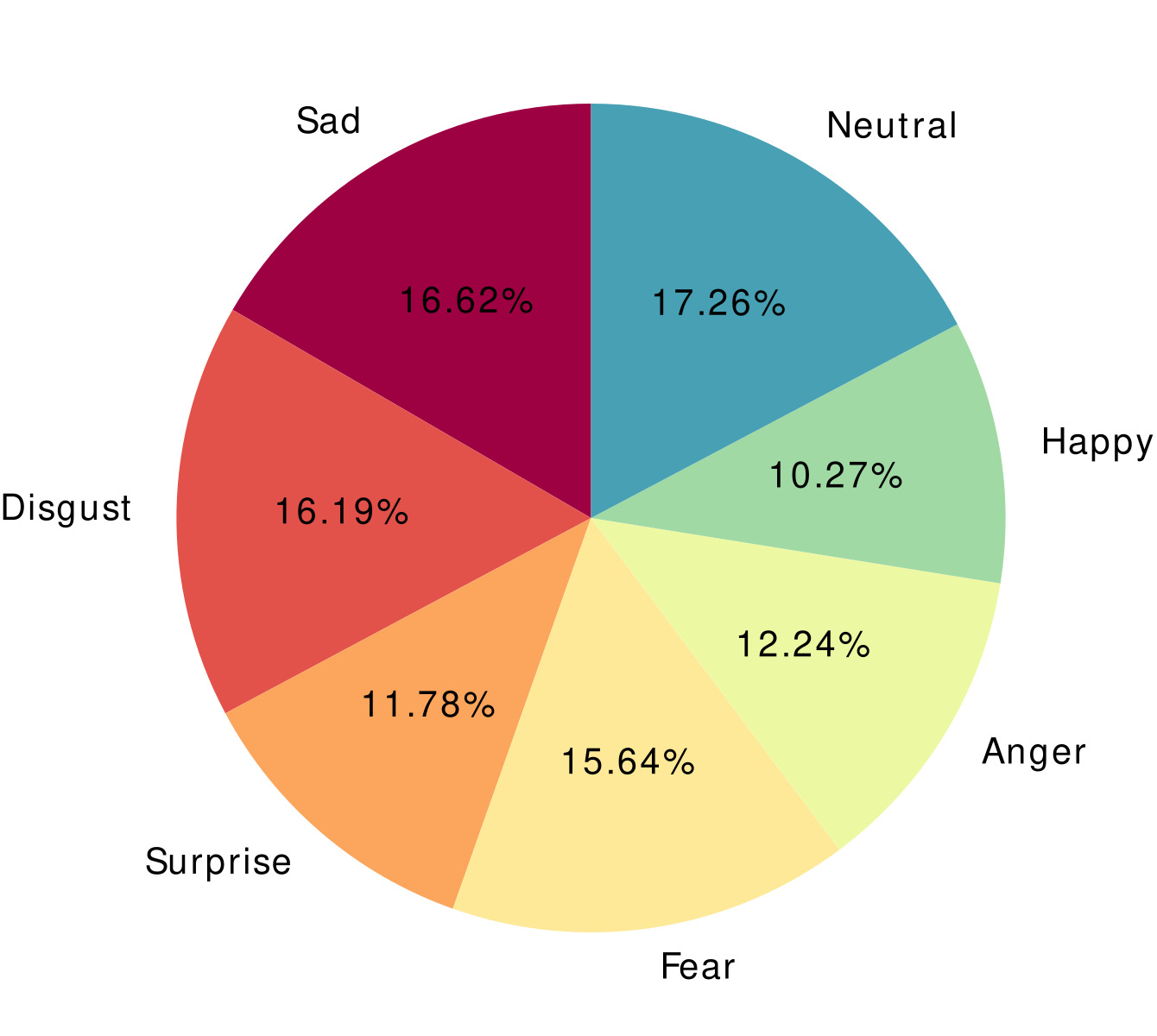}} 
    \end{tabular}
    \vspace{0ex}
    \caption{Percentage of each emotion category in the CAER and the new NCAER-S training sets}
    \label{fig:pie_chart}
\end{figure*}

Notice that the face weight and the context weight are independently computed by their corresponding networks and represent the importance of these branches. We let the two networks competitively determine which branch is more useful than the other. Then we amplify the more useful branch and lower the effect of the other by multiplying the extracted features with the corresponding weight:
\begin{align}
\mathbf{v}_f \leftarrow \mathbf{v}_f \odot w_f && \mathbf{v}_c \leftarrow \mathbf{v}_c \odot w_c
\end{align}
Finally, we use these vectors to estimate the emotion category. Specifically, in our experiments, after multiplying both $\mathbf{v}_f$ and $\mathbf{v}_c$ by their corresponding weights, we concatenate them together as the input for a network to make final predictions. Fig. \ref{fig:fusion} shows our fusion procedure in detail.

%% file: 4_experiments.tex
\section{Experiments}
\label{sec:experiments}
\subsection{Datasets}
\textbf{CAER-S} In this work, we only focus on static images with background context as our input. Therefore, we choose the static CAER (CAER-S) dataset \cite{caer} to validate our method. The CAER-S dataset contains 70K static images extracted from a total of 13201 video clips of 79 TV shows. Each image is labeled with one of seven universal emotions: \sloppy \texttt{anger}, \texttt{fear}, \texttt{disgust}, \texttt{happiness}, \texttt{neutral}, \texttt{sadness} and \texttt{surprise}. We follow the standard split proposed by \cite{caer} for training, validation and testing, respectively.

\textbf{Novel CAER-S} (NCAER-S) While experimenting with the CAER-S dataset, we observe that there is a correlation between images in the training and the test sets, which can make the model less robust to changes in data and may not generalize well on unseen samples. More specifically, many images in the training and the test set of the CAER-S dataset are extracted from the same video, hence making them look very similar to each other. To cope with this issue, we propose a novel way to extract static frames from the CAER video clips to create a new static image dataset called Novel CAER-S (NCAER-S). In particular, frames extracted from the training, validation, and test sets of the CAER dataset are separately put into the corresponding training, validation, and test sets of the new NCAER-S dataset. In particular, for each video in the original CAER dataset, we split the video into multiple parts, each part is approximately $2$s long. Then we randomly select one frame of each part to include in the new NCAER-S dataset. Any original video that provides frames for the training set will be removed from the testing set. This process assures the new dataset is novel while the training frames and testing frames are never from one original input video.



With our selection method, we ensure that images in the validation and test sets are independent of those in the training set. We also make sure that the numbers of extracted frames of each emotion category are approximately equal to tackle the imbalance problem of the CAER dataset and prevent bias towards prominent emotions.


The statistics of the original CAER and the new NCAER-S training sets are shown in Fig. \ref{fig:pie_chart} and Table \ref{table:datasets}. The new split NCAER-S dataset can be downloaded at \url{https://bit.ly/NCAERS_dataset}.

\begin{table}
	\centering
    \caption{The number of images in each emotion category in the NCAER-S training set}
    \vspace{1ex}
	\begin{tabular}{lc}
		\toprule
		Emotion &  Number of images \\ 
		\midrule
        \texttt{Angry}  & 2272 \\
        \texttt{Disgust}  & 3004\\
        \texttt{Fear}  & 2902 \\
        \texttt{Happy}  & 1905 \\ 
        \texttt{Neutral} & 3202 \\ 
        \texttt{Sad} & 3084 \\ 
        \texttt{Surprise} & 2186 \\
        \midrule
        Total &  18555 \\
		\bottomrule
	\end{tabular}
    \label{table:datasets}
\end{table}

\subsection{Experimental Setup}
\textbf{Evaluation Metric}. Classification accuracy is the standard evaluation metric that is widely used to measure the reliability of automated emotion recognition systems in the literature \cite{deepfer_survey}\cite{caer}\cite{ck_plus}\cite{afew}\cite{affectnet}. To compare our results with previous approaches quantitatively, as in \cite{deepfer_survey}\cite{caer} we use the overall classification accuracy as the evaluation metric:
\begin{equation}
    \text{Accuracy} = \frac{1}{N}{\Sigma_{i=1}^N\mathds{1}\{\hat{y}_i=y_i\}}
\end{equation}
where $\mathds{1}$ is the indicator function, $N$ is the total number of samples in the dataset, $\hat{y}_i$ and $y_i$ is the network prediction and ground-truth category of the $i$-th example, respectively.

\textbf{Baselines}. We compare the results of our proposed \textbf{G}lobal-\textbf{L}ocal \textbf{A}ttention for E\textbf{mo}tion \textbf{R}ecognition network (GLAMOR-Net) with the following methods as baselines: AlexNet \cite{alexnet}, VGGNet \cite{vgg}, ResNet \cite{resnet}, CAER-Net-S\cite{caer}. The results of AlexNet, VGGNet, and ResNet on the CAER-S dataset are reported in two cases: using the ImageNet dataset as the pre-trained model, and fine-tuning these networks on this dataset. Note that, these results are taken from \cite{caer} paper. On the CAER-S, we also compare our method to several recent state-of-the-art approaches. GRERN {\cite{GRERN}} utilized a multi-layer Graph Convolutional Network (GCN) to exploit the relationship among different regions in the context. EfficientFace {\cite{EfficientFace}} proposed an efficient lightweight network and utilized the label distribution to handle the ambiguity of real-world emotions. MA-Net {\cite{manet}} designed a highly complicated architecture based on ensemble learning of multiple regions to handle the occlusion and pose variation problems. We report the results of our GLAMOR-Net with two different backbones: the original encoding module introduced in section \ref{section:encoding_module} and ResNet-18 \cite{resnet}.

\textbf{Implementation Details}. Our networks are implemented using Tensorflow 2.0 framework \cite{tf}. For optimization, we use the SGD optimization algorithm and standard cross-entropy loss function:
\begin{equation}
    \mathcal{L} = -\frac{1}{N}\sum_{i=1}^N \log p_i^{(y_i)}
\end{equation}
where $p_i^{(y_i)}$ is the predicted probability for the true emotion category $y_i $ of the $i$-th sample and $N$ is the total number of samples in the dataset.

Given an input image, we first use the CNN based face detector in the dlib library \cite{dlib} to detect the face coordinates. The detected face is then cropped and resized to 96 x 96 and fed to the Facial Encoding Module. To create input for the Context Encoding Module, we mask the facial region in the original image and resize it to 128 x 171, then we apply random crop during the training phase and center crop during the inference phase to the final size of 112 x 112. We use a dropout layer before the final layer with a dropout rate of 0.5 to reduce the effect of overfitting. During training, we observe that the fusion network is very unstable and easily affected by random factors. Specifically, the weights of the face branch or the context branch in the Fusion Module can easily take a value near 0 or 1, which means the model completely ignores information extracted from one of the branches. To tackle this problem, we first train the Facial Encoding Module and the Context Encoding Module separately, then jointly train both modules and the fusion network in an end-to-end manner.

\subsection{Results}
\subsubsection{Results on the CAER-S dataset}
Table \ref{table:CAERSresult} summarizes the results of our network and other recent state-of-the-art methods on the CAER-S dataset \cite{caer}. This table clearly shows that integrating our GLA module can significantly improve the accuracy performance of the recent CAER-Net. In particular, our GLAMOR-Net (original) achieves 77.90\% accuracy, which is a +4.38\% improvement over the CAER-Net-S. When compared with other recent state-of-the-art approaches, the table clearly demonstrates that our GLAMOR-Net (ResNet-18) outperforms all those methods and achieves a new state-of-the-art performance with an accuracy of 89.88\%. This result confirms our global-local attention mechanism can effectively encode both facial information and context information to improve the human emotion classification results. 

\begin{figure}[t]
    \centering
    \includegraphics[width=0.45\textwidth]{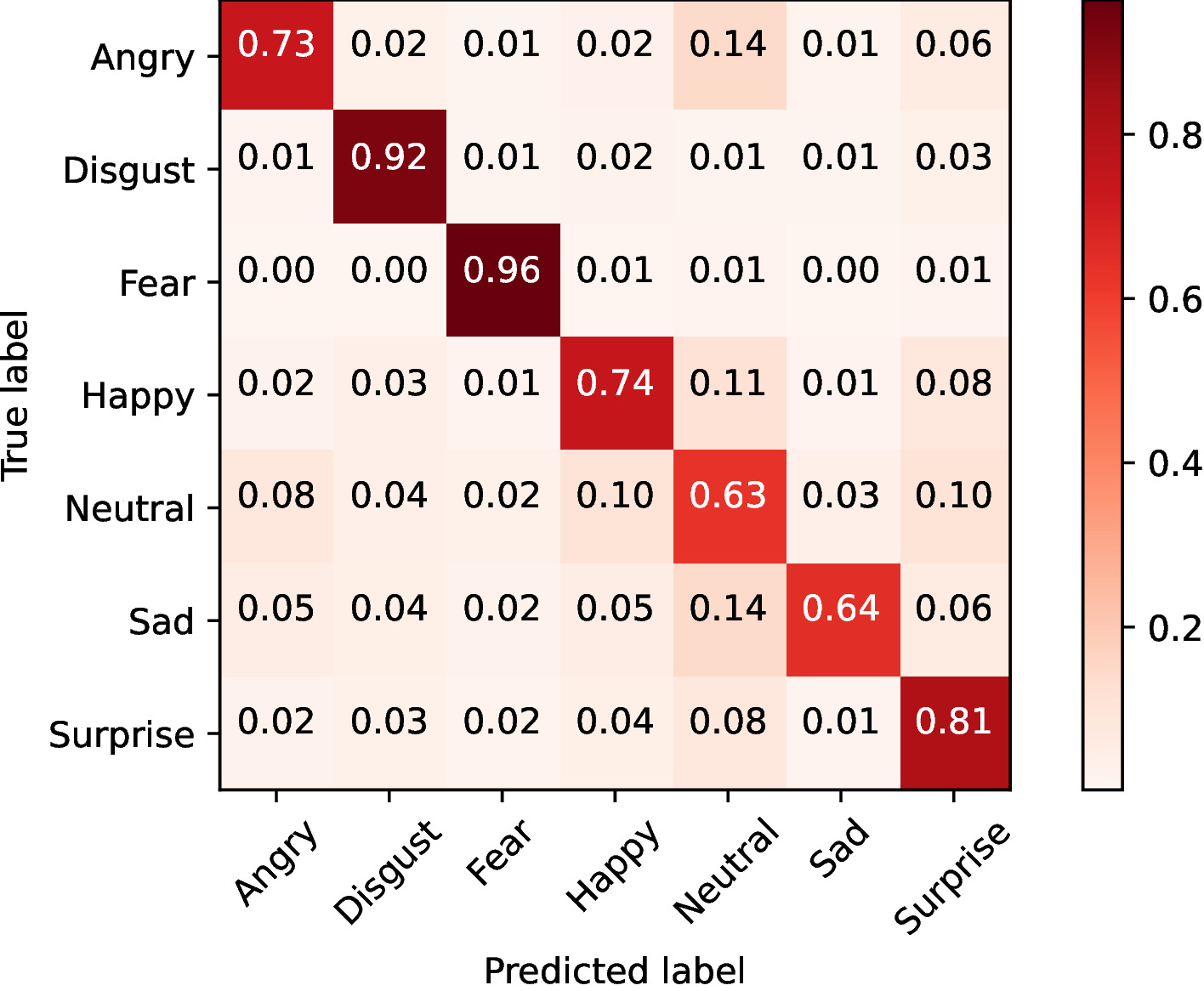}
    \vspace{2ex}
    \caption{The confusion matrix of our GLAMOR-Net (original) results on the CAER-S test set}
    \label{fig:caers_confusion_matrix}
\end{figure}

Fig \ref{fig:caers_confusion_matrix} shows the confusion matrix of the GLAMOR-Net (original) on the CAER-S dataset. Overall, the model achieves the highest accuracy on the \texttt{fear} class with $0.96$ accuracy. The \texttt{neutral} class has the lowest accuracy of $0.63$ as there are many misclassifications from other classes.

\begin{table}
    \caption{Classification accuracy of baseline methods and our GLAMOR-Net on the CAER-S dataset}
    \vspace{1ex}
	\label{table:CAERSresult}
	\centering
	\begin{tabular}{lll}
		\toprule
		Methods & Year & Accuracy (\%) \\ 
		\midrule
        ImageNet-AlexNet \cite{alexnet} & 2012 & 47.36\\
        ImageNet-VGGNet \cite{vgg} & 2015 & 49.89\\
        ImageNet-ResNet \cite{resnet} & 2016 &57.33 \\
        \midrule
        Fine-tuned AlexNet \cite{alexnet} & 2012 & 61.73\\
        Fine-tuned VGGNet \cite{vgg} & 2015 & 64.85 \\
        Fine-tuned ResNet \cite{resnet} & 2016 & 68.46 \\
        CAER-Net-S \cite{caer} & 2019 & 73.52 \\ 
        \midrule
        GRERN \cite{GRERN} & 2020 & 81.31 \\
        EfficientFace \cite{EfficientFace} & 2021 &  81.48\\
        MA-Net \cite{manet}  & 2021 & 88.42 \\
        \midrule
        GLAMOR-Net (original)  & 2021 & \textbf{77.90}\\
        GLAMOR-Net (ResNet-18)  & 2021 & \textbf{89.88} \\
		\bottomrule
	\end{tabular}
\end{table}

\subsubsection{Results on the NCAER-S dataset} 
On the NCAER-S dataset, we compare our results with three recent methods: VGG16 \cite{vgg}, ResNet50 \cite{resnet}, and CAER-Net-S \cite{caer}. The results from the VGG16 and ResNet50 models are reproduced as baseline methods. We finetune the VGG16 and the ResNet50 from the pre-trained models on VGG-Face and ImageNet, respectively. Our GLAMOR-Net (original) and CAER-Net-S are trained from scratch for a fair comparison.

\begin{table}
	\centering
		\caption{Classification accuracy of baseline methods and our GLAMOR-Net on the NCAER-S dataset}
		\vspace{1ex}
	\begin{tabular}{ll}
		\toprule
		Methods & Accuracy (\%) \\ 
		\midrule
        VGG16 \cite{vgg} & 42.85 \\ 
        ResNet50 \cite{resnet} & 41.41 \\ 
        CAER-Net-S \cite{caer} & 44.14 \\
        \midrule
        GLAMOR-Net (original) & \textbf{46.91} \\ 
		\bottomrule
	\end{tabular}
    \label{table:CAERSresplittedresult}
\end{table}

\begin{figure}[]
    \centering
    \includegraphics[width=0.45\textwidth]{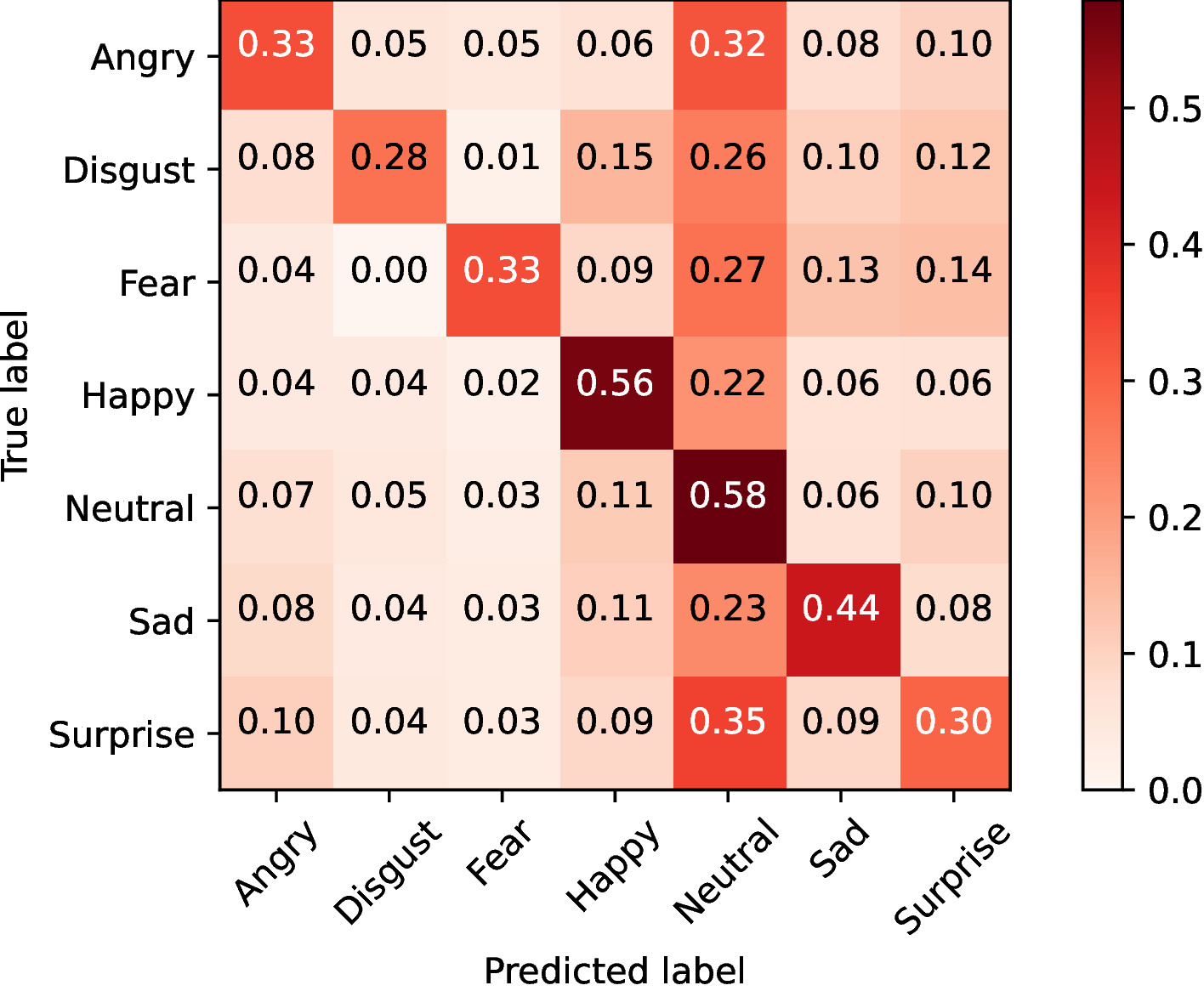}
    \caption{The confusion matrix of our GLAMOR-Net (original) results on the NCAER-S test set}
    \label{fig:ncaers_confusion_matrix}
\end{figure}

Table \ref{table:CAERSresplittedresult} reports the comparative results of our GLAMOR-Net and other recent methods.
This table shows that the GLAMOR-Net architecture outperforms all other architectures and achieves the highest performance. In particular, our network increases classification accuracy by 2.77\% compared to the second-highest model CAER-Net-S. These results also validate the effectiveness of our proposed global-local attention mechanism integrated into the GLAMOR-Net. We note that the result of VGG16 pre-trained on VGG-Face is surprisingly better than the result of ResNet50 pre-trained on ImageNet dataset. This is explainable as the pre-trained weight on VGG-Face carries more meaningful information than the pre-trained weight on ImageNet, which includes many non-face images.

\begin{figure*}
\captionsetup[subfigure]{labelformat=empty, farskip=1.5pt}
\centering
\setlength\tabcolsep{1.5pt} 
\begin{tabular}{cccccc}
\subfloat{\includegraphics[width = 0.15\textwidth, height=0.15\textwidth]{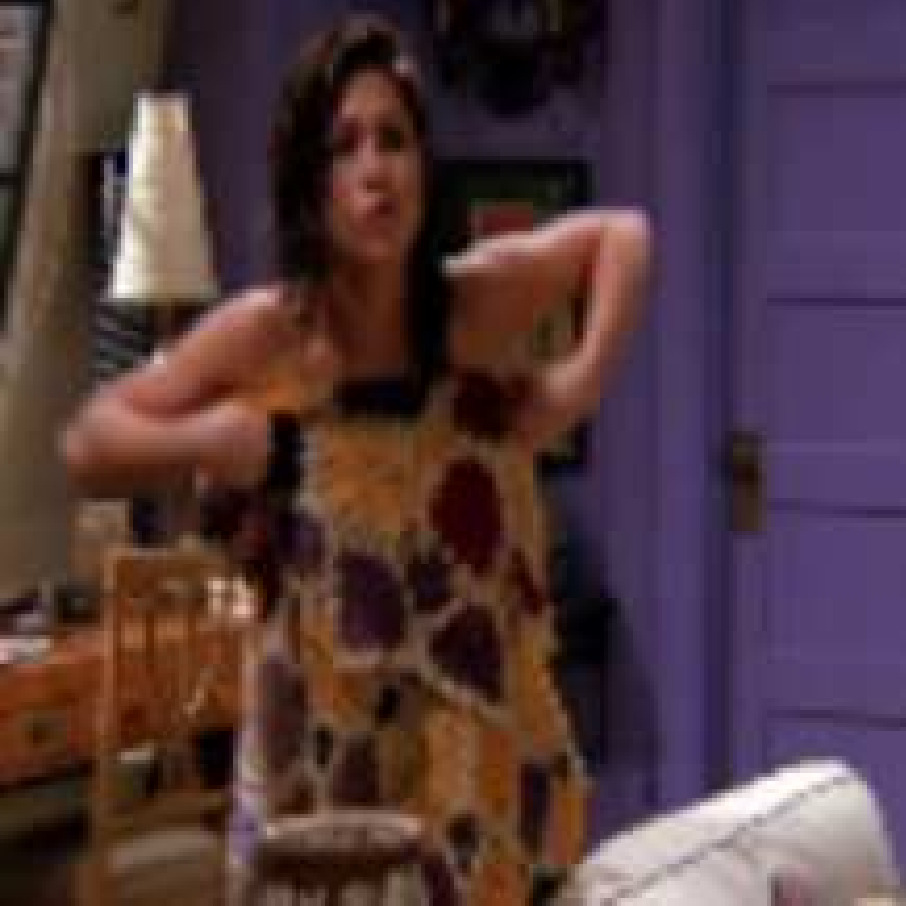}} &
\subfloat{\includegraphics[width = 0.15\textwidth]{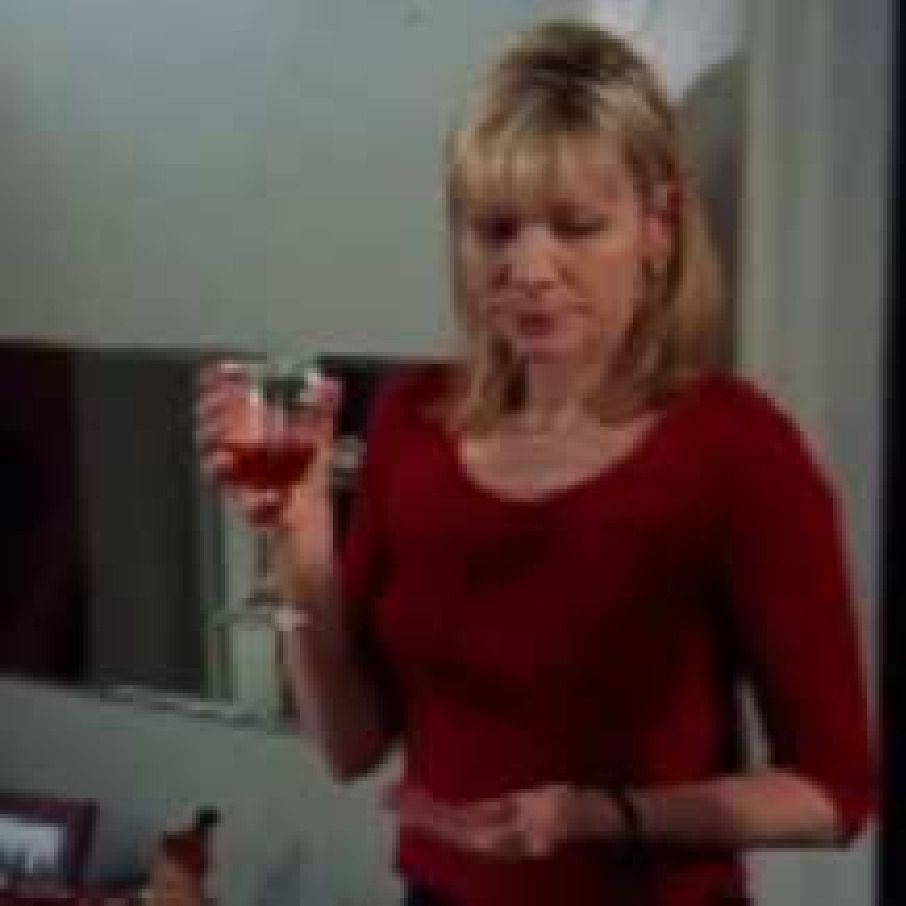}} &
\subfloat{\includegraphics[width = 0.15\textwidth]{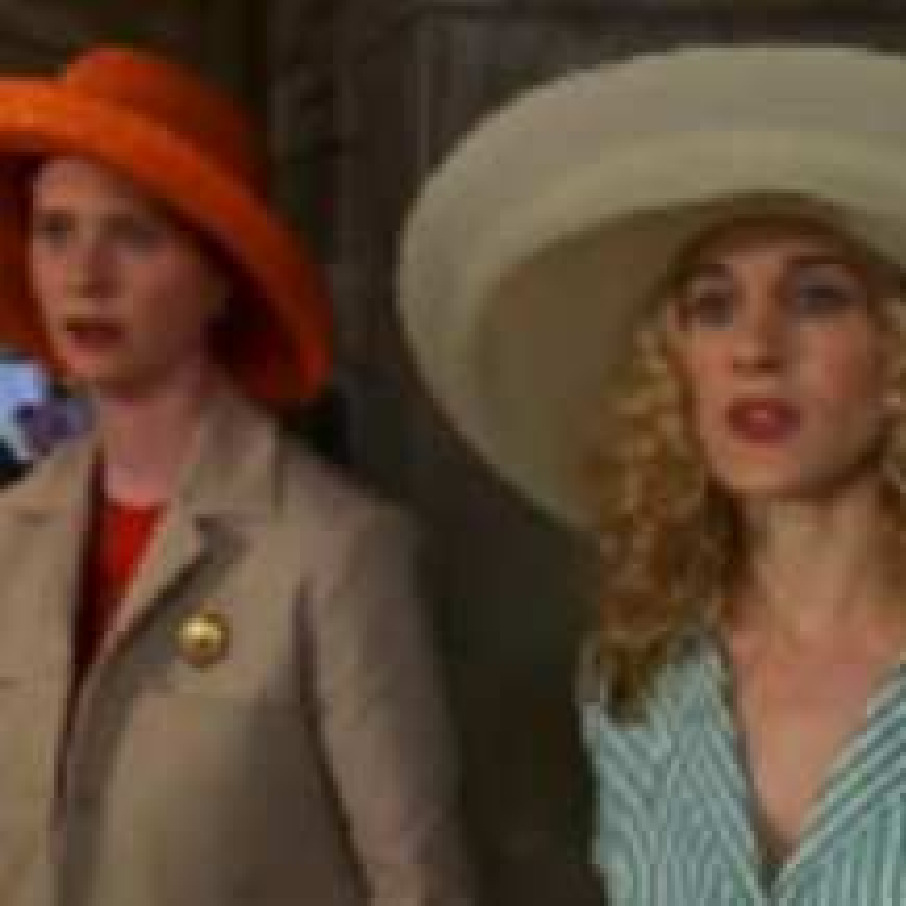}} &
\subfloat{\includegraphics[width = 0.15\textwidth]{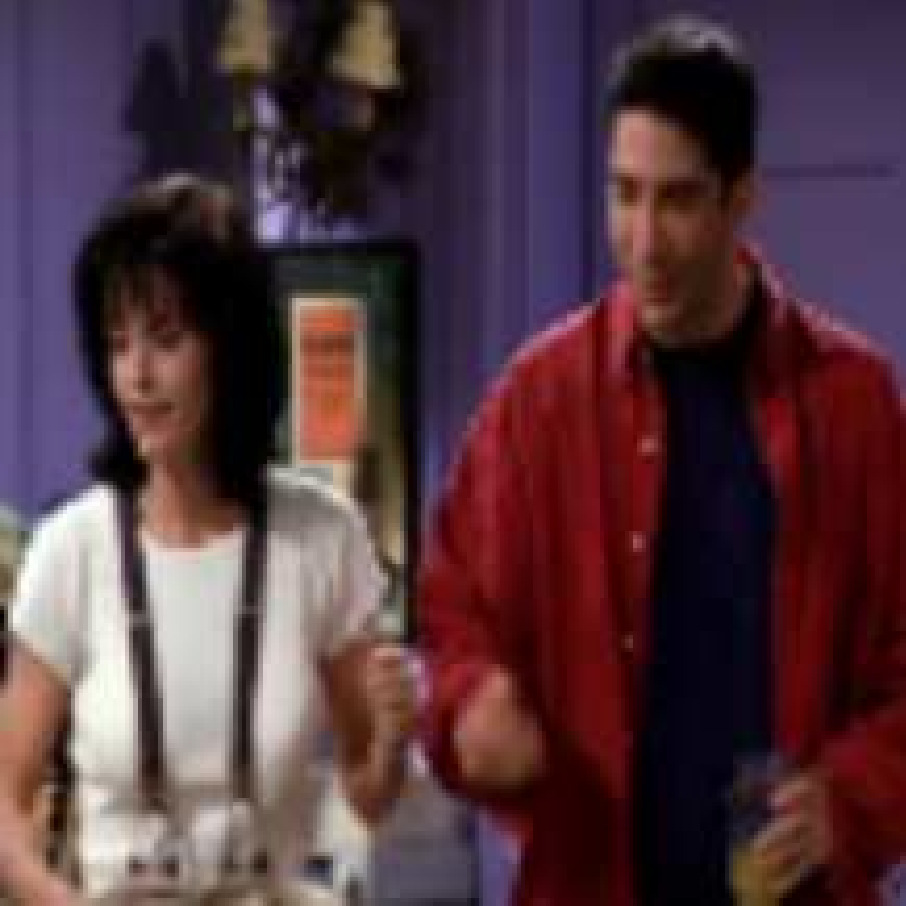}} &
\subfloat{\includegraphics[width = 0.15\textwidth]{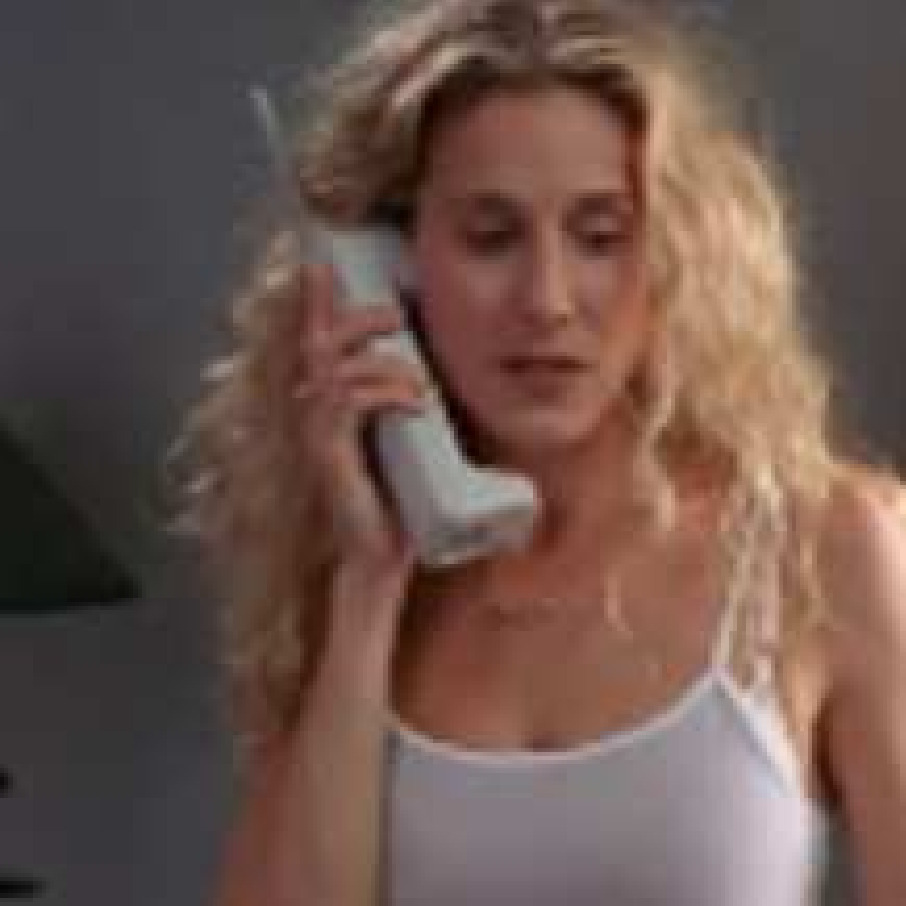}} &
\subfloat{\includegraphics[width = 0.15\textwidth]{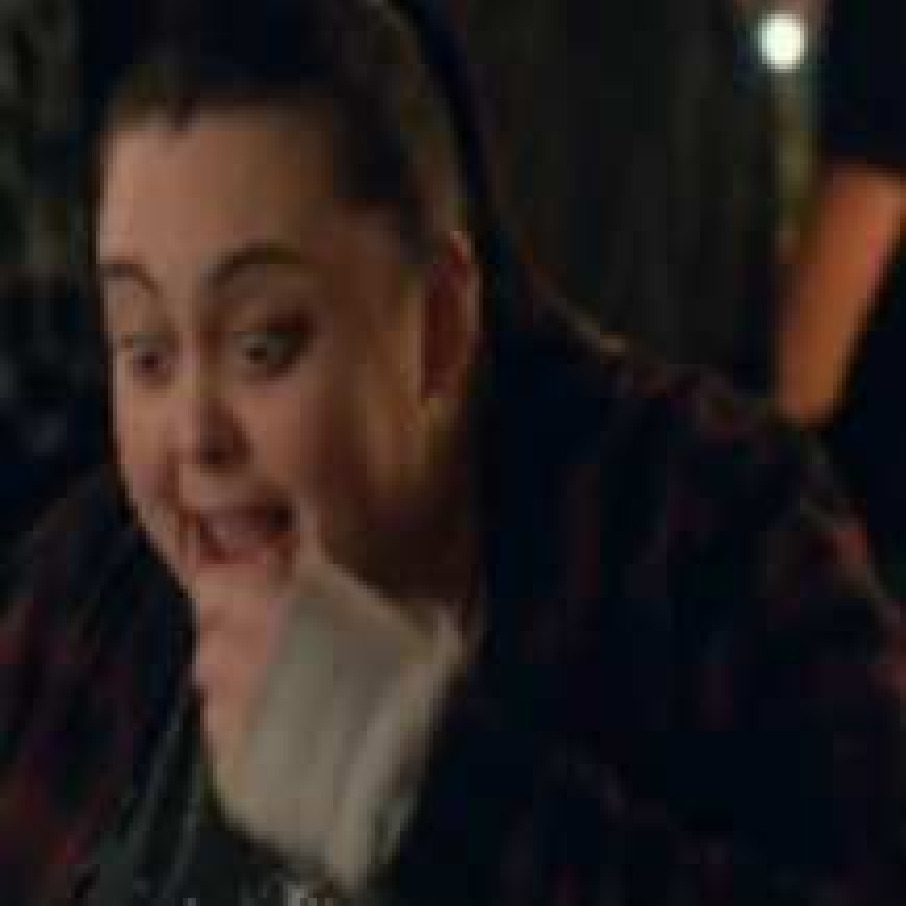}} \\
\subfloat{\includegraphics[width = 0.15\textwidth]{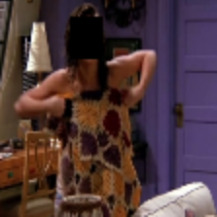}} &
\subfloat{\includegraphics[width = 0.15\textwidth]{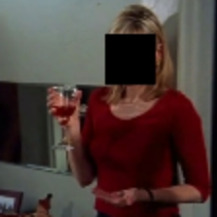}} &
\subfloat{\includegraphics[width = 0.15\textwidth]{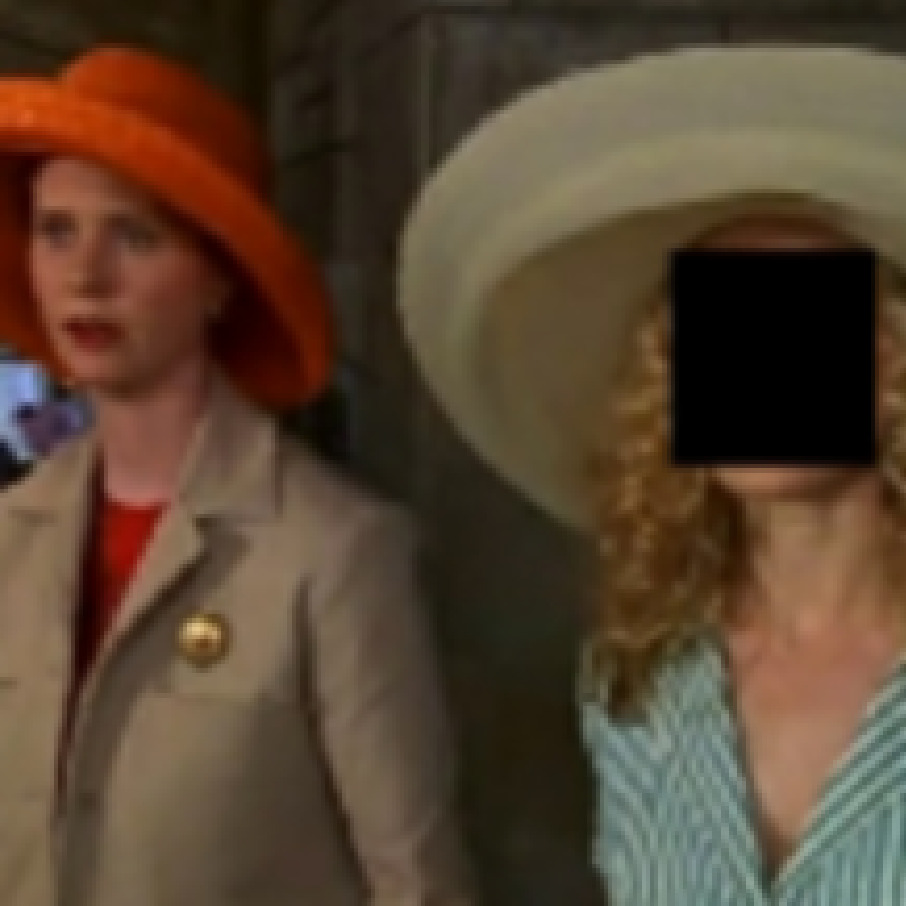}} &
\subfloat{\includegraphics[width = 0.15\textwidth]{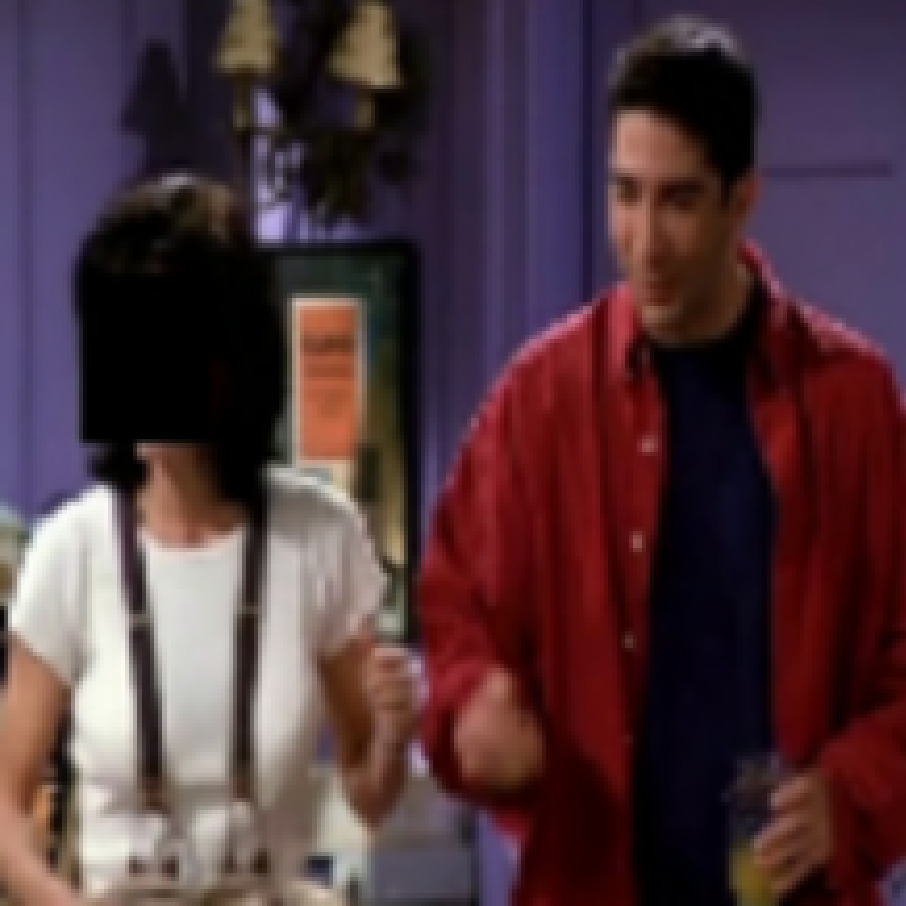}} &
\subfloat{\includegraphics[width = 0.15\textwidth]{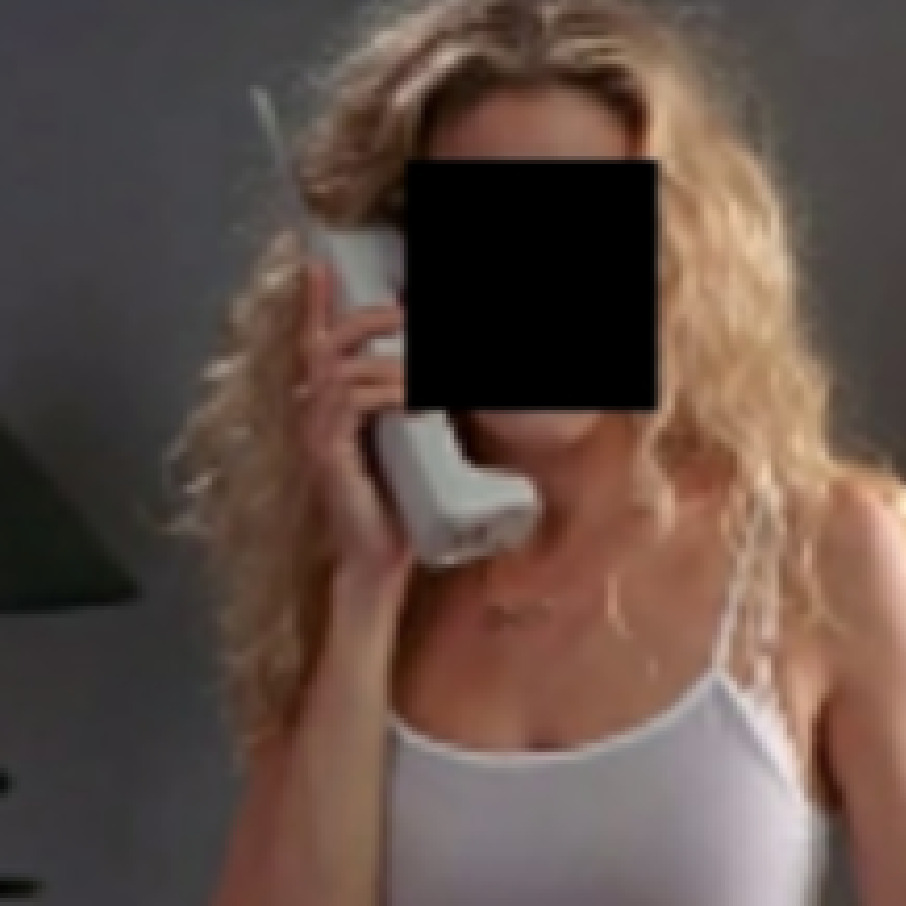}} &
\subfloat{\includegraphics[width = 0.15\textwidth]{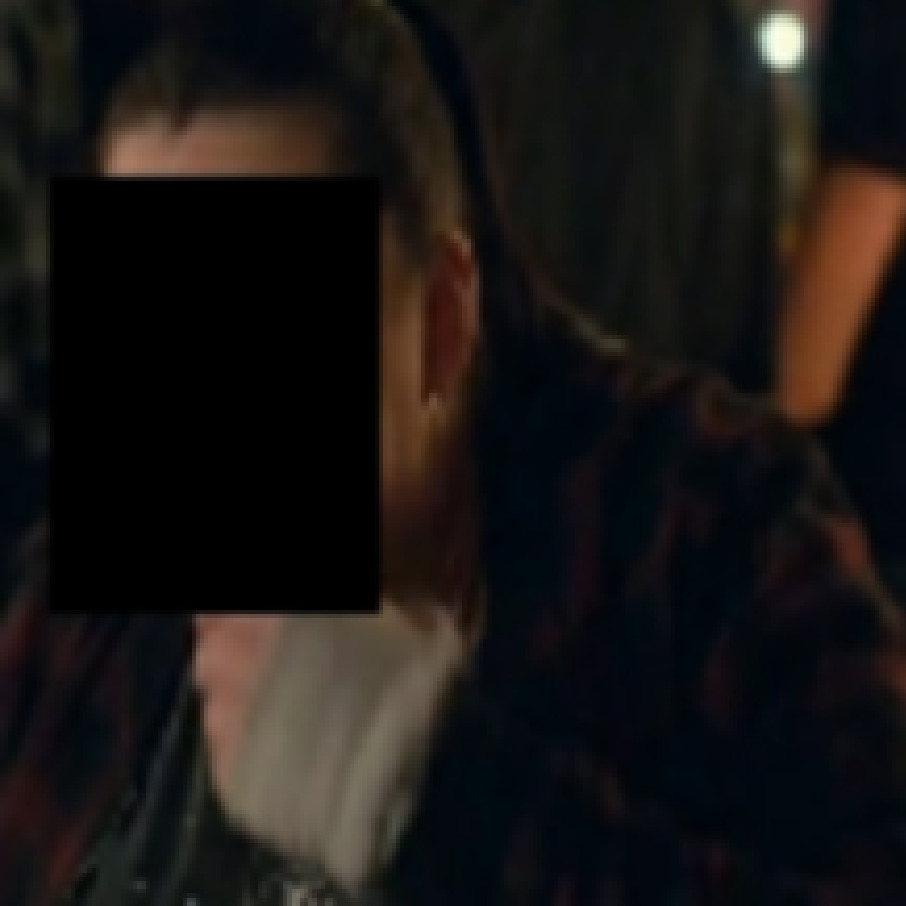}} \\
\subfloat{\includegraphics[width = 0.15\textwidth]{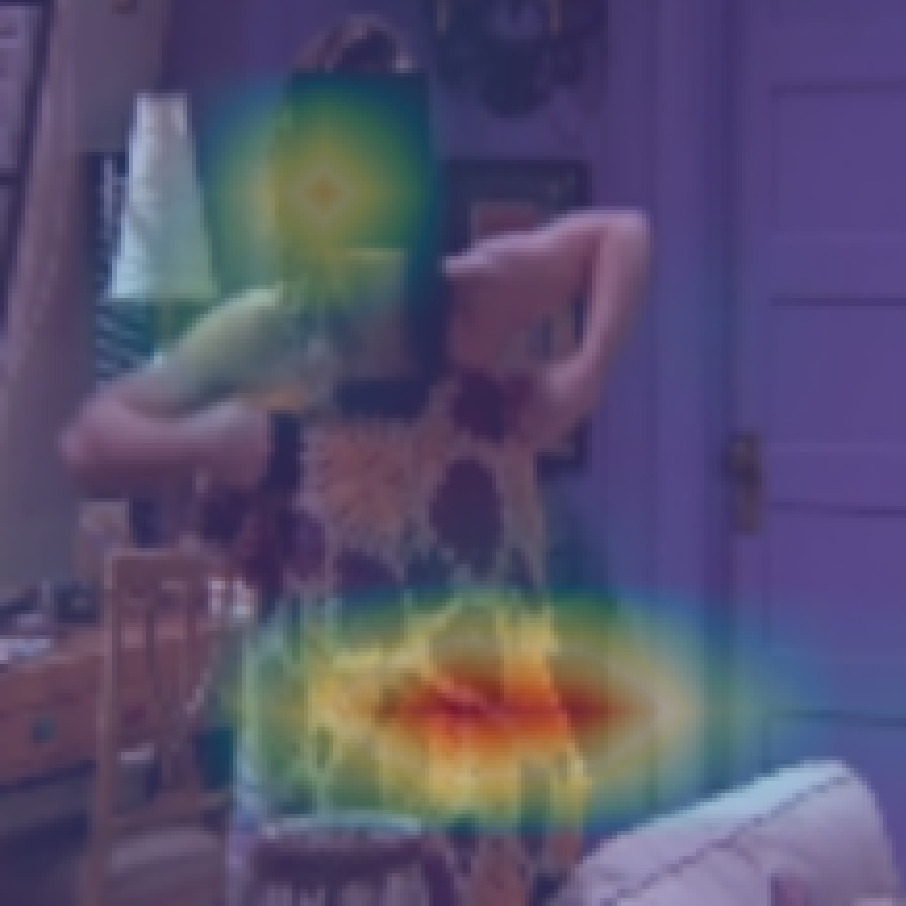}} &
\subfloat{\includegraphics[width = 0.15\textwidth]{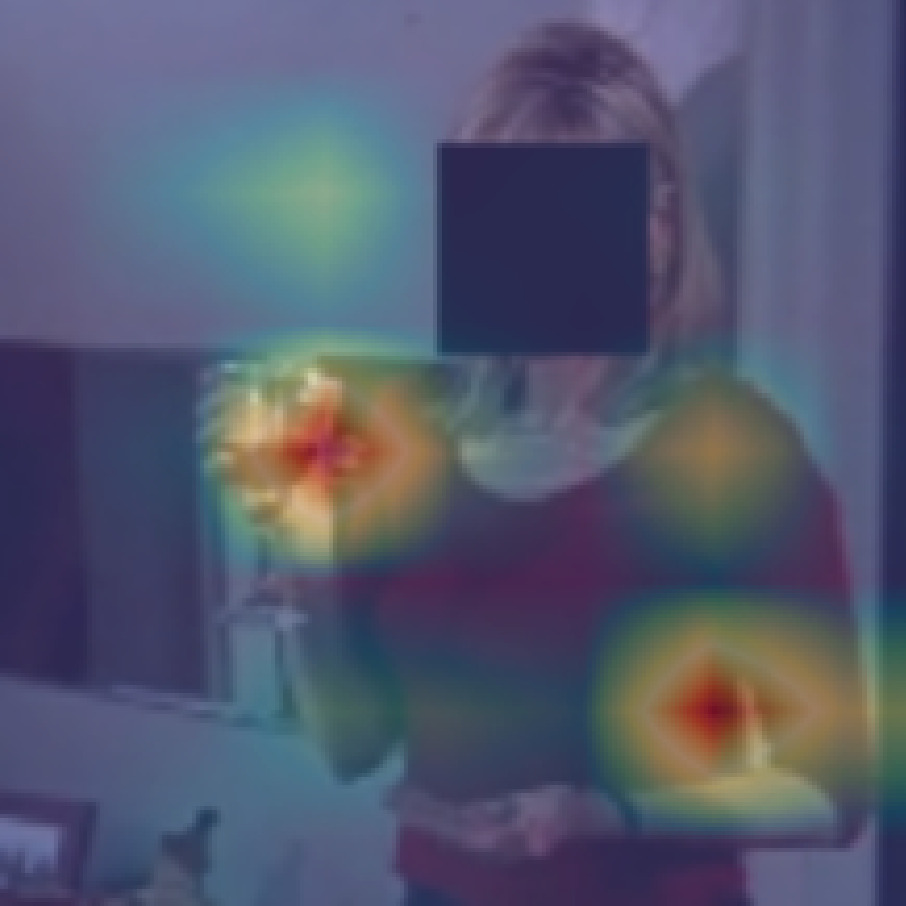}} &
\subfloat{\includegraphics[width = 0.15\textwidth]{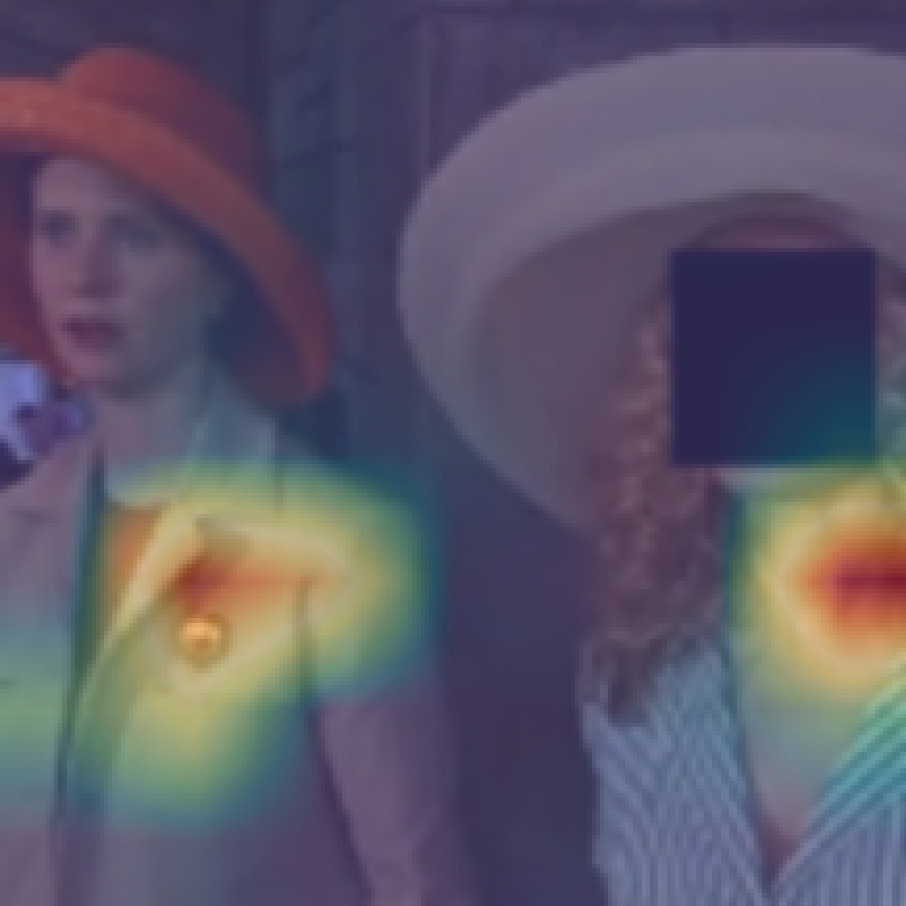}} &
\subfloat{\includegraphics[width = 0.15\textwidth]{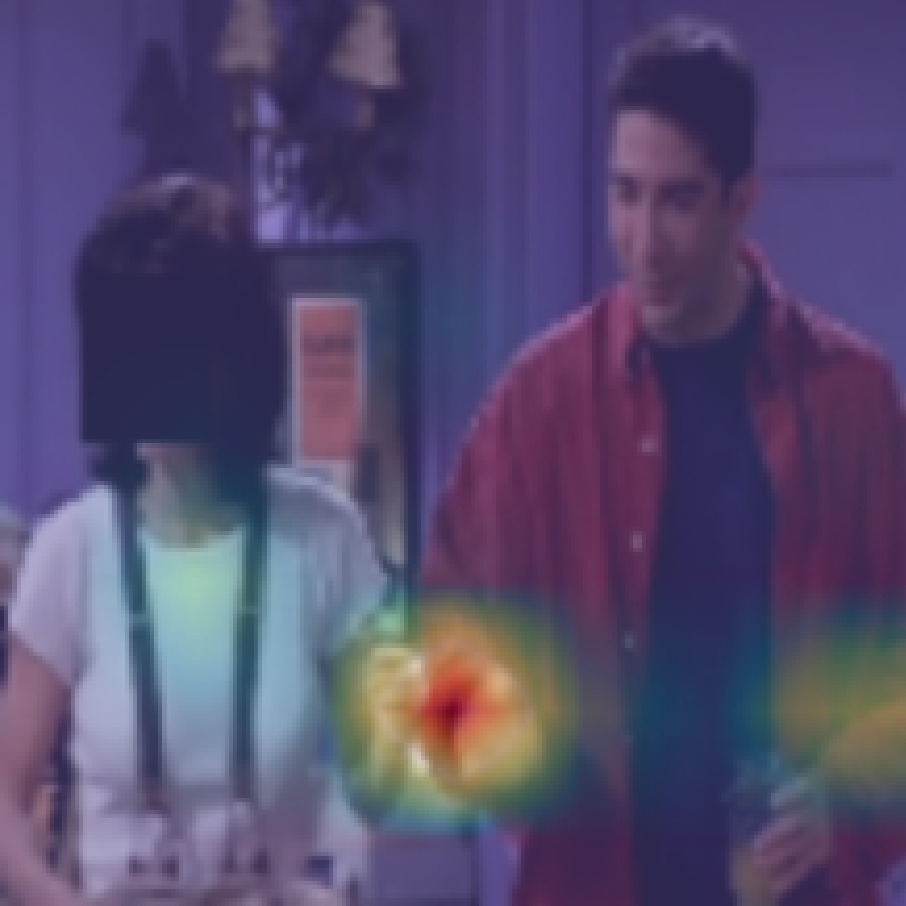}} &
\subfloat{\includegraphics[width = 0.15\textwidth]{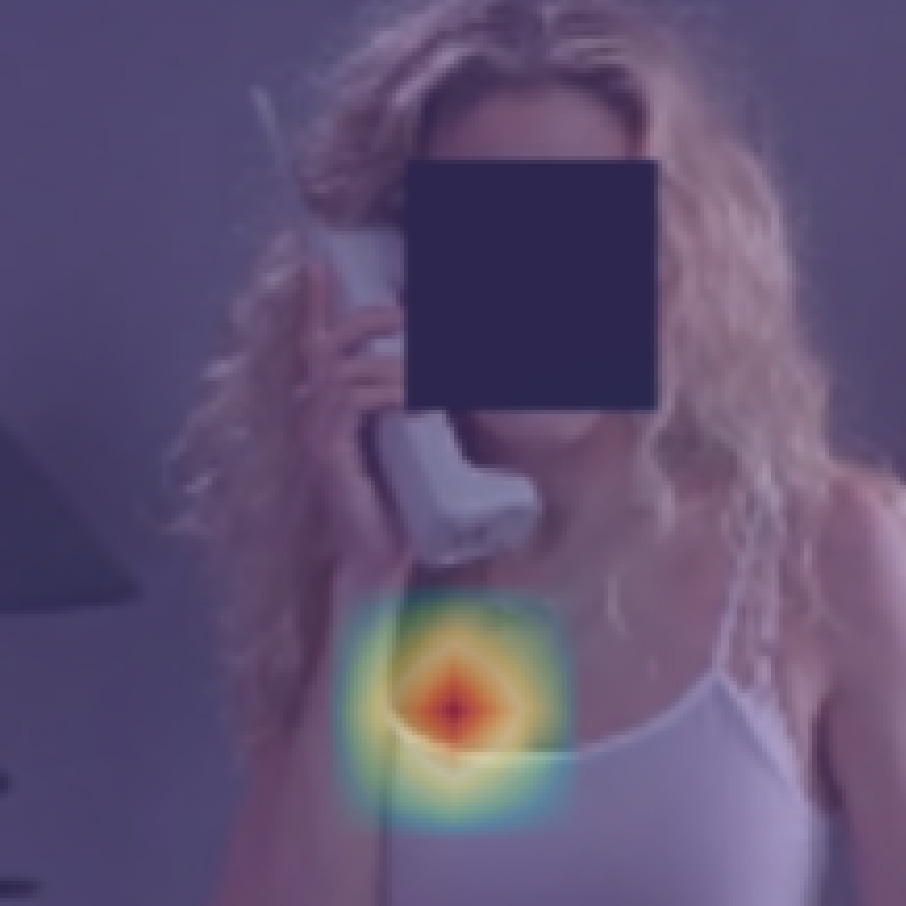}} &
\subfloat{\includegraphics[width = 0.15\textwidth]{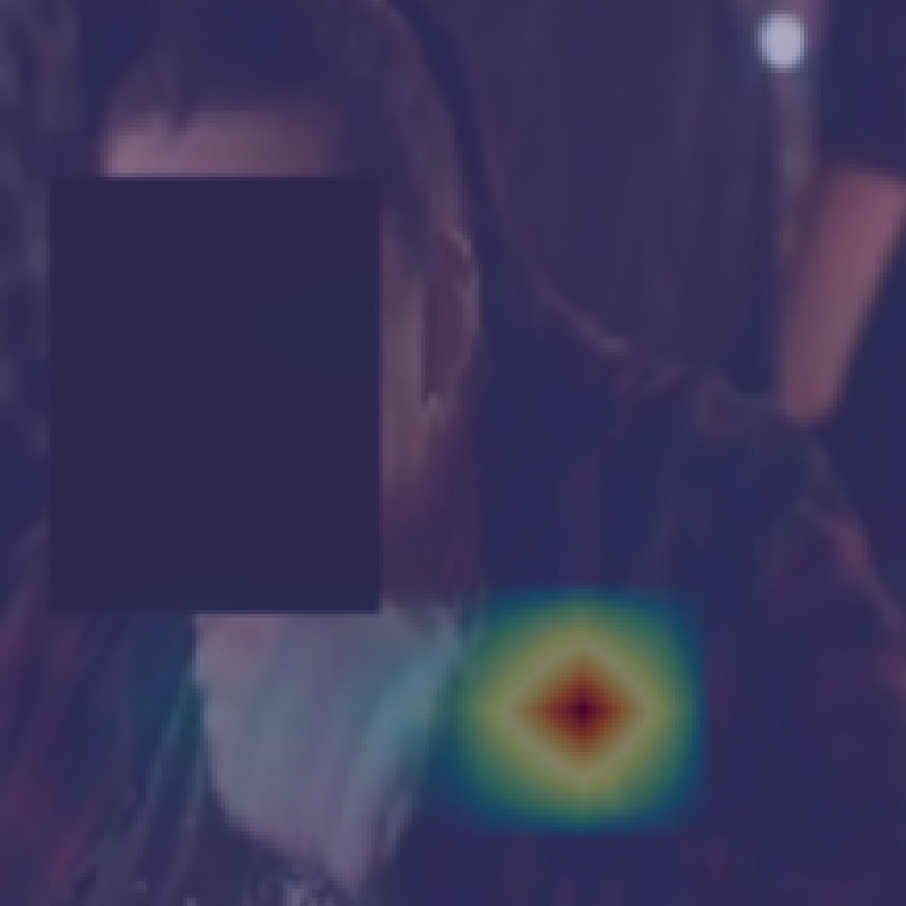}}\\
\subfloat[(a) Anger]{\includegraphics[width = 0.15\textwidth]{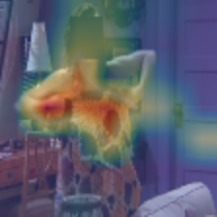}} &
\subfloat[(b) Disgust]{\includegraphics[width = 0.15\textwidth]{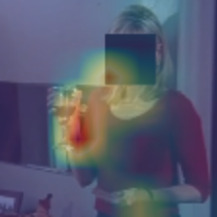}} &
\subfloat[(c) Fear]{\includegraphics[width = 0.15\textwidth]{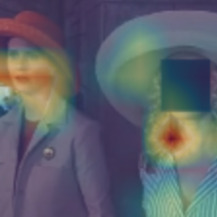}} &
\subfloat[(d) Happy]{\includegraphics[width = 0.15\textwidth]{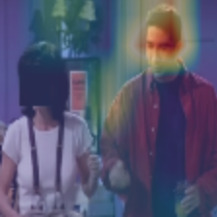}} &
\subfloat[(e) Sad]{\includegraphics[width = 0.15\textwidth]{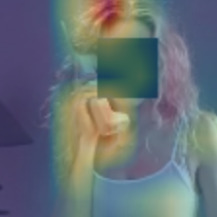}} &
\subfloat[(f) Surprise]{\includegraphics[width = 0.15\textwidth]{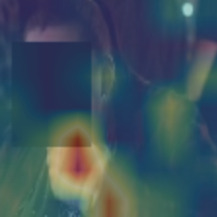}}
\end{tabular}
\vspace{0ex}
\caption{Visualization of the attention maps. From top to bottom: original image in the NCAER-S dataset, image with masked face, attention map of the CAER-Net-S, and attention map of our GLAMOR-Net}
\label{fig:attention_map}
\end{figure*}

Also from Table \ref{table:CAERSresplittedresult}, we can see that the classification accuracy of the models is much lower than those in Table \ref{table:CAERSresult}. The reason behind this is the new NCAER-S is more challenging than the original CAER-S dataset. As mentioned earlier, to construct the NCAER-S dataset, we eliminate the correlation between the train and the test samples as much as we can. Specifically, we separately resample image frames from clips of the train and test sets of the CAER dataset to mitigate the train and test dependency. Moreover, note that the size of the new dataset is only less than one-third of the original one, which also limits the amount of information that the models can exploit. However, our GLAMOR-Net still consistently outperforms other state-of-the-art methods despite the challenges of the NCAER-S dataset and shows competitive results.

The confusion matrix of our GLAMOR-Net evaluated on the NCAER-S dataset is given in Fig \ref{fig:ncaers_confusion_matrix}. The two categories with the highest accuracy are \texttt{happy} and \texttt{neutral} while the \texttt{disgust} emotion has the lowest accuracy of 0.28.
It can also be inferred from the confusion matrix that our model mostly confuses \texttt{neutral} with other emotion categories as most of the misclassified examples of the six categories: \texttt{angry}, \texttt{disgust}, \texttt{fear}, \texttt{happy}, \texttt{sad} and \texttt{surprise} fall into the class \texttt{neutral}. It may be because the facial emotion in the NCAER-S dataset is weakly expressed, which makes it more difficult to identify and distinguish other emotions from the \texttt{neutral} class. 

In summary, we can conclude that our method consistently improves the results on both the original CAER-S and the challenging NCAER-S datasets. Note that although we follow the same procedure as in \cite{caer}, our proposed Global-Local Attention Module is the key difference that helps enhance the accuracy of the emotion recognition task. The results reported in Table \ref{table:CAERSresult} and Table \ref{table:CAERSresplittedresult} verify that with the assistance of our attention strategy, the classification accuracy is significantly improved. We believe that if a more sophisticated neural architecture is adopted, the performance will be further boosted.
\subsubsection{Analysis}
\begin{table}
	\centering
    \caption{Ablation study of our proposed method on the NCAER-S dataset. `w/F', `w/mC', `w/fC', `w/CA', `w/GLA' denote using the output of the Facial Encoding Module, the Context Encoding Module with masked faces as input, the Context Encoding Module with visible faces as input, the standard Context Attention in CAER-Net-S \cite{caer} and our Global-Local Attention Module, respectively, as input to the Fusion Network}
    \vspace{1ex}
	\resizebox{0.48\textwidth}{!}{\begin{tabular}{ccccccl}
		\toprule
		Settings & w/F & w/mC & w/fC & w/CA & w/GLA & Accuracy (\%) \\ 
		\midrule
        (\textit{i})& \cmark & & & & & 42.58 \\ 
        \hline
        \multirow{3}{*}{(\textit{ii})} & & \cmark & & & & 41.18 \\ 
        & & \cmark & & \cmark & & 41.27 \\ 
        & & \cmark & & & \cmark & 42.24 \\ 
        \hline 
        \multirow{2}{*}{(\textit{iii})} & & & \cmark & \cmark & & 41.94 \\ 
        & & & \cmark & & \cmark & 42.66 \\ 
        \hline
        \multirow{3}{*}{(\textit{iv})} & \cmark & \cmark & & & & 43.19 \\ 
        & \cmark & \cmark & & \cmark & & 44.14 \\ 
        & \cmark & \cmark & & & \cmark & \textbf{46.91} \\         
		\bottomrule
	\end{tabular}}
    \label{table:ablation_study}
\end{table}

\begin{table}
	\centering
		\caption{$p$ value of the Stuart-Maxwell test for each pair of methods that are used in the setting (\textit{iv}) of Table \ref{table:ablation_study}. Each element on the main diagonal is the test result of the agreement between the model prediction and the observed data (ground-truth label).}
		\vspace{1ex}
	\begin{tabular}{llll}
		\toprule
		Methods & w/GLA & w/CA & w/o Attention\\ 
		\midrule
        w/GLA  & $3.0 \times 10^{-2}$ & $1.69 \times 10^{-14}$ & $1.38 \times 10^{-53}$ \\
        w/CA & & $1.33 \times 10^{-10}$ & $3.33 \times 10^{-14}$\\ 
        w/o Attention  & & & $2.81 \times 10^{-38}$ \\
 
		\bottomrule
	\end{tabular}
    \label{table:significance_test}
\end{table}

To further analyze the contribution of each component in our proposed method, we experiment with 4 different input settings on the NCAER-S dataset: (\textit{i}) face only, (\textit{ii}) context only with the facial region being masked, (\textit{iii}) context only with the facial region visible, and (\textit{iv}) both face and context (with masked face). When the context information is used, we compare the performance of the model with different context attention approaches (no attention, standard attention module in CAER-Net-S and our GLA module). Note that to compute the saliency map with the proposed GLA in the (\textit{ii}) and (\textit{iii}) setting, we extract facial features using the Facial Encoding Module, however, these features are only used as the input of the GLA module to guide the context attention map learning process and \textit{not} as the input of the Fusion Network to predict the emotion category. The performances of these settings are summarized in Table \ref{table:ablation_study}. 

The results clearly show that our GLA consistently helps improve performance in all settings. Specifically, in setting (\textit{ii}), using our GLA achieves an improvement of 1.06\% over method without attention, 0.97\% over standard attention module in CAER-Net-S \cite{caer}. It is also noteworthy that when the context with visible faces is utilized as in setting (\textit{iii}), using the attention module in the CAER-Net-S achieves 41.94\% accuracy, lower than the one using only the cropped face in setting (\textit{i}) by 0.64\%, while using our GLA module achieves higher accuracy (42.66\% vs. 42.58\%). Our GLA also improves the performance of the model when both facial and context information is used to predict emotion. Specifically, our model with GLA achieves the best result with an accuracy of 46.91\%, which is higher than the method with no attention 3.72\% and standard attention module in \cite{caer} 2.77\%. The results from Table \ref{table:ablation_study} show the effectiveness of our Global-Local Attention module for the task of emotion recognition. They also verify that the use of both the local face region and global context information is essential for improving emotion recognition accuracy.

In order to emphasize the contribution of the Attention module to the final results, we conduct Stuart-Maxwell test for each pair of methods that are used in the setting (\textit{iv}) of Table \ref{table:ablation_study}.The Stuart-Maxwell test is the generalized version of McNemar test \cite{mcnemar_test} which is generally used for testing the significant difference of multi-class classification models. The resulted $p$-values of the tests are shown in Table \ref{table:significance_test}. Note that the lower $p$ value indicates stronger statistical disagreement between the two compared methods. Overall, we can see that all of the models have significant different error rates. Furthermore, the higher value on the main diagonal would imply stronger agreement between the model prediction and the observed data, which means the performance is better. In conjunction with the results in Table \ref{table:ablation_study}, we can statistically confirm that our GLA module performs better than other attention mechanisms.

\subsubsection{Fusion methods comparison}

\begin{table}
	\centering
		\caption{Results of different fusion strategies on the NCAER-S dataset.}
		\vspace{1ex}
	\begin{tabular}{ll}
		\toprule
		Methods & Accuracy (\%) \\ 
		\midrule
        
        Dubey \textit{et al.} \cite{lbpdad} & 44.33 \\
        GLAMOR-Net + \emph{Fusion Add} & 45.62 \\
        GLAMOR-Net + \emph{Fusion Max} & 46.26 \\
        GLAMOR-Net + \emph{Fusion Net} & \textbf{46.91}\\ 
		\bottomrule
	\end{tabular}
    \label{table:fusion_results}
\end{table}

To study the effectiveness of the information obtained from multiple modalities via different fusion strategies, we conduct experiment by alternatively changing the Fusion Module with multiple Fusion operators while keeping other components of the system unchanged. Specifically, the Element-wise Addition (\emph{Fusion Add}), Element-wise Maximum (\emph{Fusion Max)}) and our \emph{Fusion Net} are studied in our experiment. Furthermore, we also compare our method with recent work by Dubey \textit{et al.} \cite{lbpdad}. Table \ref{table:fusion_results} summarizes the results from our experiment. As shown in this table, the performance of our network using \emph{Fusion Net} is superior to other fusion strategies. However, we notice that the results from other fusion techniques are also very competitive. This shows that the fusion strategy is also an important module in the emotion recognition task, however the final result is also affected by the extracted features from the feature extraction and attention modules.



\subsubsection{{Backbone architectures}}
\begin{table}
	\centering
		\caption{Accuracy of different encoding network architectures}
		\vspace{1ex}
	\resizebox{0.48\textwidth}{!}{\begin{tabular}{l|l|r|l|l}
		\toprule
		Method & Backbone &  \#Params & CAER-S & NCAER-S \\ 
		\midrule
        GLAMOR-Net & Original &  2.23M & 77.90 & 46.91 \\
        GLAMOR-Net & MobileNetV2 {\cite{mobilenetv2}} &  5.83M & 85.44 & 47.52 \\ 
        GLAMOR-Net & ResNet-18 {\cite{resnet}} &  22.90M & \textbf{89.88} & \textbf{48.40} \\
		\bottomrule
	\end{tabular}}
    \label{table:ablation_backbone}
\end{table}
{We further study the effect of different Encoding network architectures. Specifically, the MobileNetV2 {\cite{mobilenetv2}} and ResNet-18 {\cite{resnet}} are adopted as the backbone network to extract features for both face and context branches in our study. We use the output of the last convolutional layer as the represented feature maps. These feature maps are then fed into the GLA module and processed as in Section {\ref{sec:approach}}. We summarize the total amount of network parameters and the classification results on CAER-S and NCAER-S in Table {\ref{table:ablation_backbone}}. We observe that the ResNet-18 significantly outperforms other shallower architectures (Original and MobileNetV2) and yields the best performance with 89.88\% and 48.40\% accuracy on CAER-S and NCAER-S. However, using such complex model resulted in more memory footprint as well as computational cost. Additionally, the MobileNetV2 can balance the trade-off between accuracy and the speed of the model, which is a considerable option for deploying in environments with limited resources such as mobile devices.}



\begin{figure*}
\captionsetup[subfigure]{labelformat=empty, farskip=1.5pt}
\centering
\setlength\tabcolsep{1.5pt} 
\begin{tabular}{cccccccc}
(\textit{i}) &
\subfloat{\includegraphics[width = 0.13\textwidth, height=0.14\textwidth]{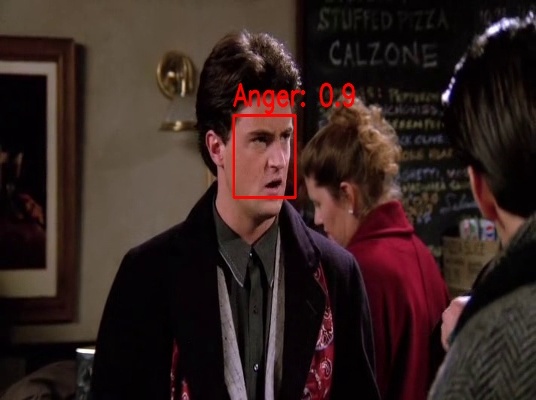}} &
\subfloat{\includegraphics[width = 0.13\textwidth, height=0.14\textwidth]{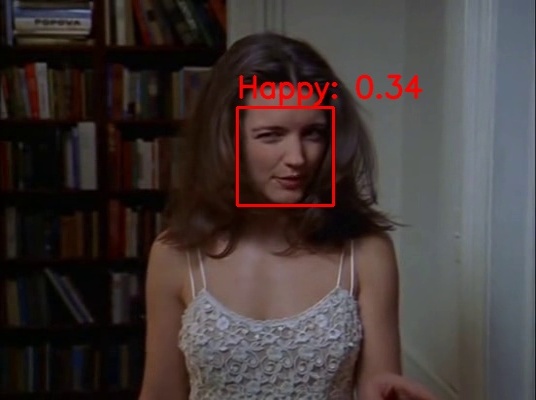}} &
\subfloat{\includegraphics[width = 0.13\textwidth, height=0.14\textwidth]{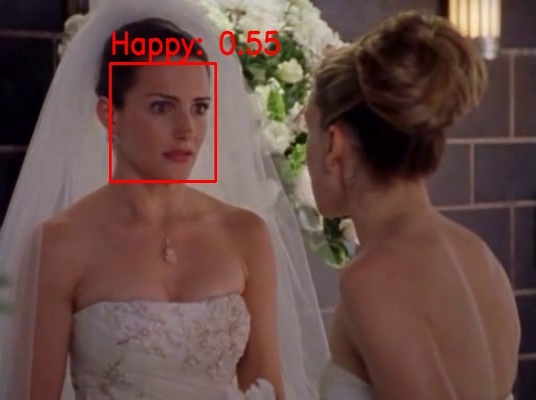}} &
\subfloat{\includegraphics[width = 0.13\textwidth, height=0.14\textwidth]{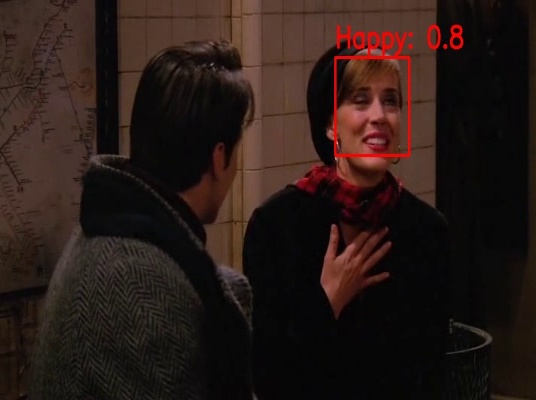}} &
\subfloat{\includegraphics[width = 0.13\textwidth, height=0.14\textwidth]{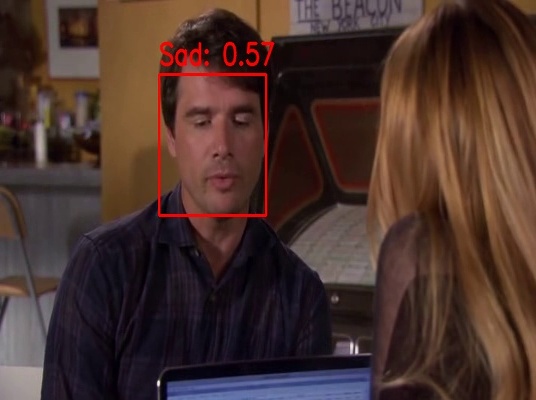}} &
\subfloat{\includegraphics[width = 0.13\textwidth, height=0.14\textwidth]{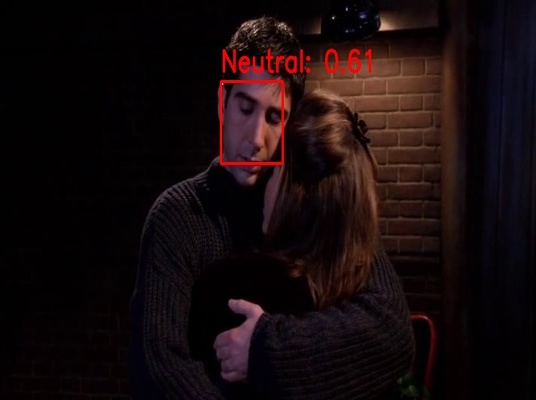}} &
\subfloat{\includegraphics[width = 0.13\textwidth, height=0.14\textwidth]{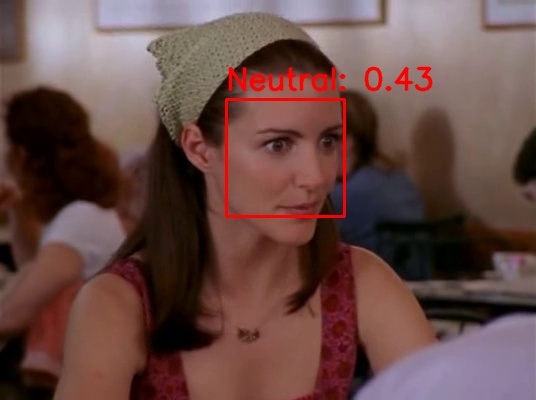}} \\
(\textit{ii}) &
\subfloat{\includegraphics[width = 0.13\textwidth, height=0.14\textwidth]{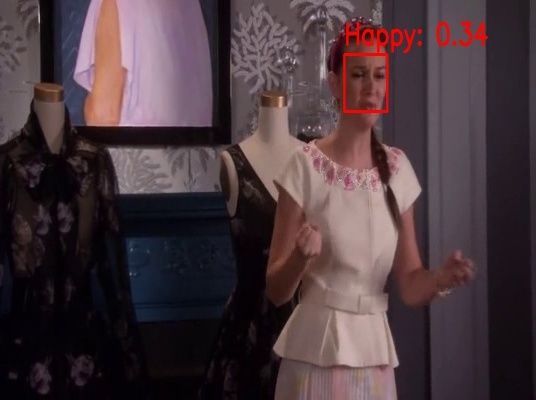}} &
\subfloat{\includegraphics[width = 0.13\textwidth, height=0.14\textwidth]{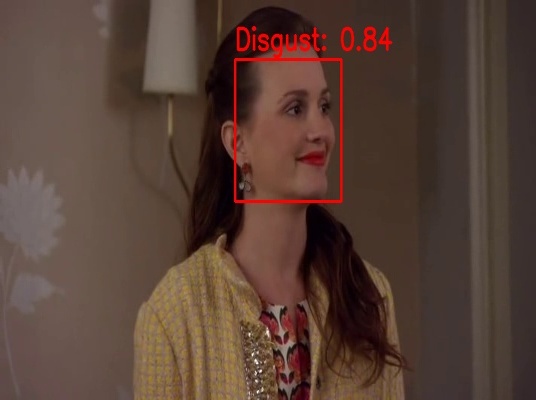}} &
\subfloat{\includegraphics[width = 0.13\textwidth, height=0.14\textwidth]{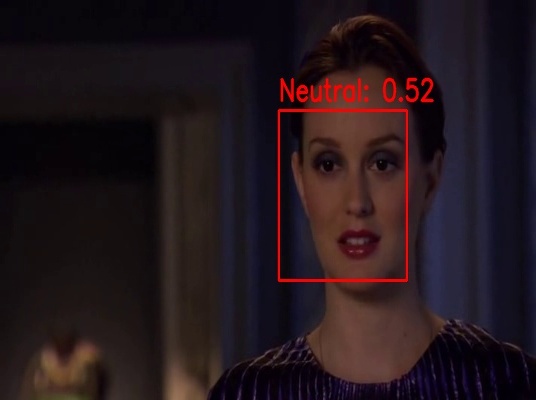}} &
\subfloat{\includegraphics[width = 0.13\textwidth, height=0.14\textwidth]{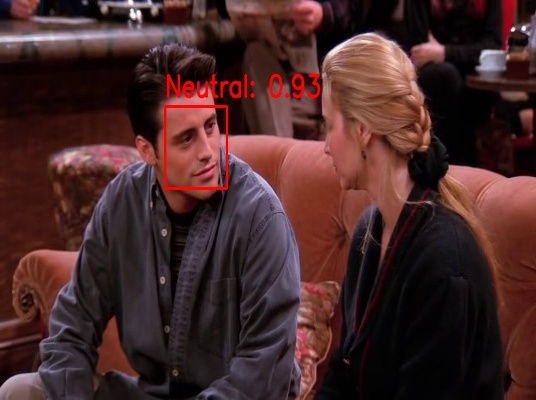}} &
\subfloat{\includegraphics[width = 0.13\textwidth, height=0.14\textwidth]{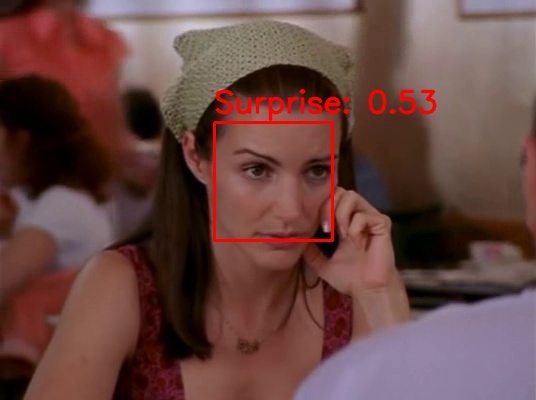}} &
\subfloat{\includegraphics[width = 0.13\textwidth, height=0.14\textwidth]{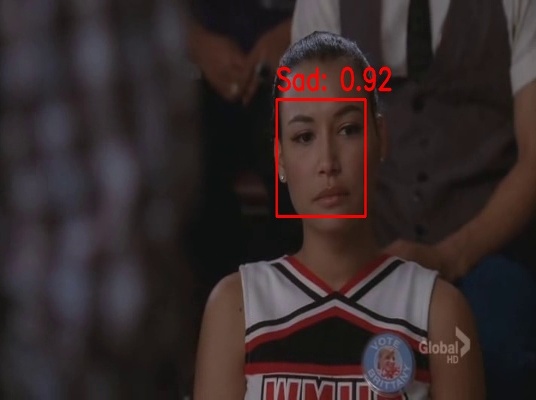}} &
\subfloat{\includegraphics[width = 0.13\textwidth, height=0.14\textwidth]{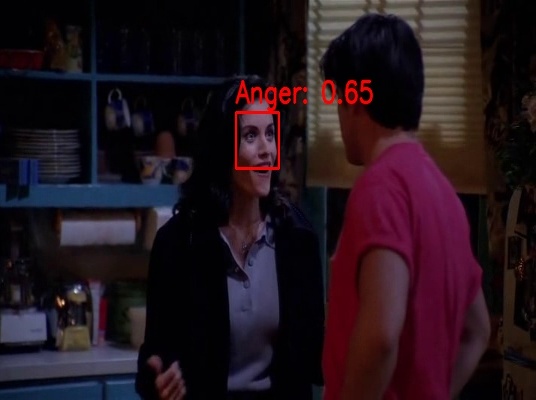}}
\\
\centering (\textit{iii}) &
\subfloat{\includegraphics[width = 0.13\textwidth, height=0.14\textwidth]{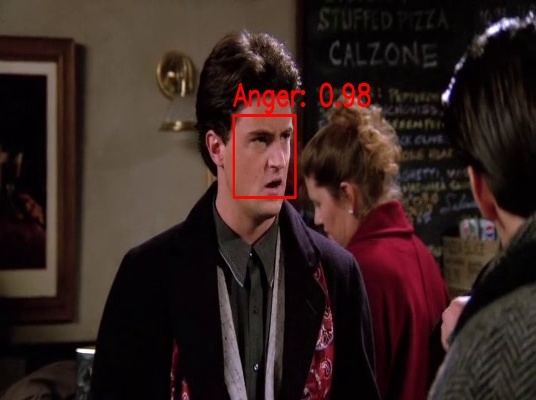}} &
\subfloat{\includegraphics[width = 0.13\textwidth, height=0.14\textwidth]{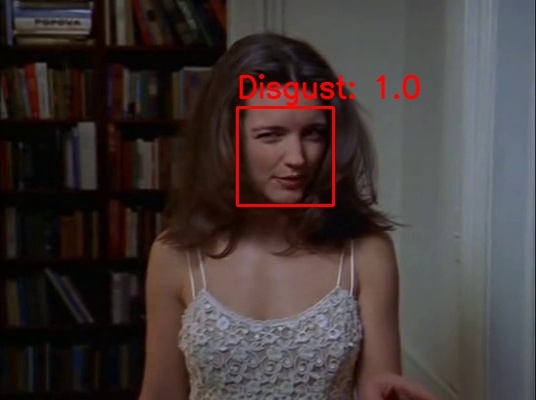}} &
\subfloat{\includegraphics[width = 0.13\textwidth, height=0.14\textwidth]{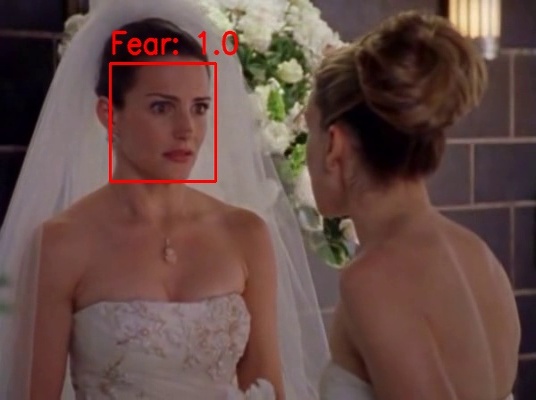}} &
\subfloat{\includegraphics[width = 0.13\textwidth, height=0.14\textwidth]{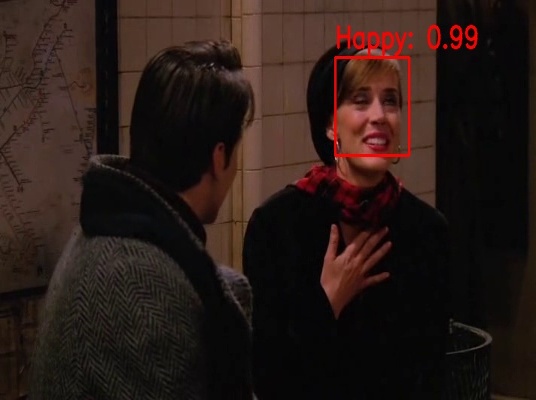}} &
\subfloat{\includegraphics[width = 0.13\textwidth, height=0.14\textwidth]{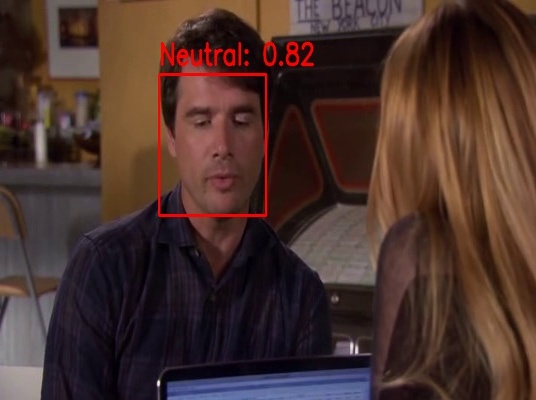}} &
\subfloat{\includegraphics[width = 0.13\textwidth, height=0.14\textwidth]{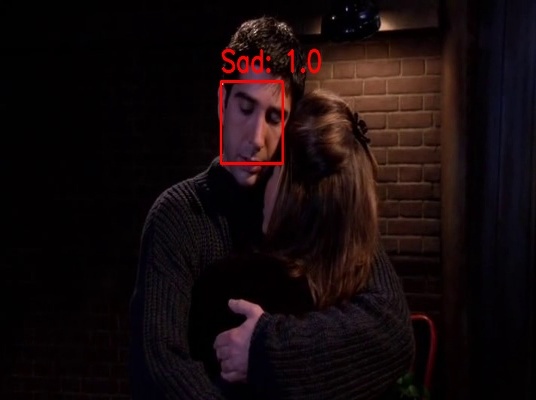}} &
\subfloat{\includegraphics[width = 0.13\textwidth, height=0.14\textwidth]{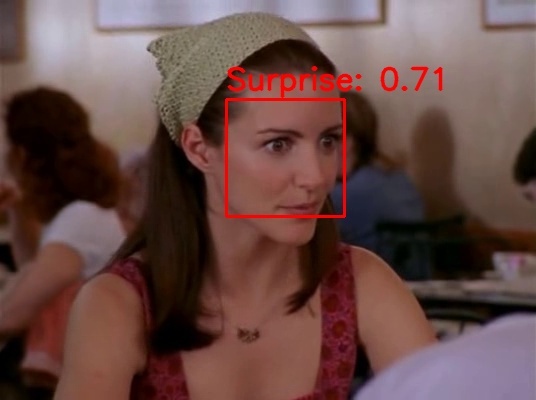}} \\
(\textit{iv}) &
\subfloat[(a) Anger]{\includegraphics[width = 0.13\textwidth, height=0.14\textwidth]{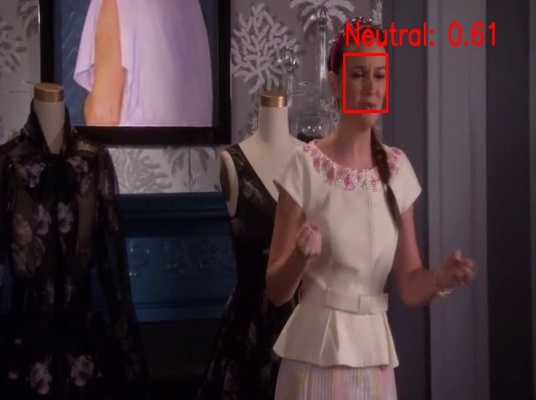}} &
\subfloat[(b) Disgust]{\includegraphics[width = 0.13\textwidth, height=0.14\textwidth]{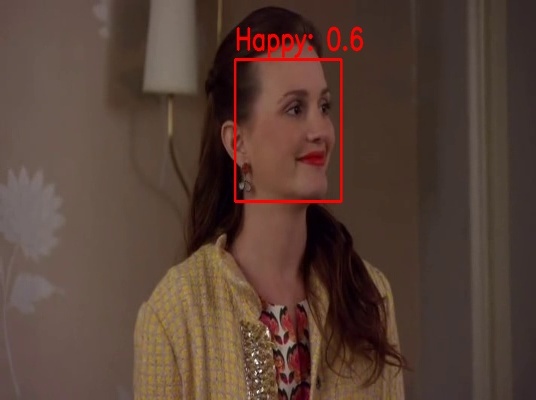}} &
\subfloat[(c) Fear]{\includegraphics[width = 0.13\textwidth, height=0.14\textwidth]{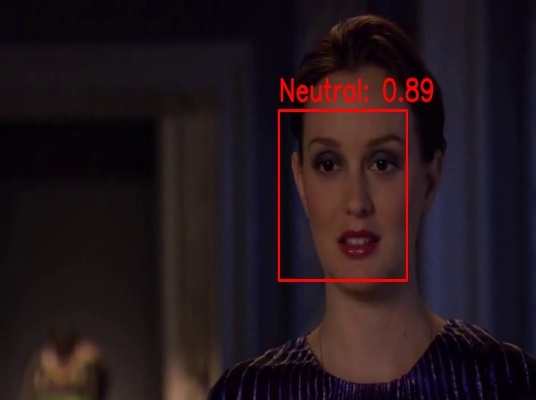}} &
\subfloat[(d) Happy]{\includegraphics[width = 0.13\textwidth, height=0.14\textwidth]{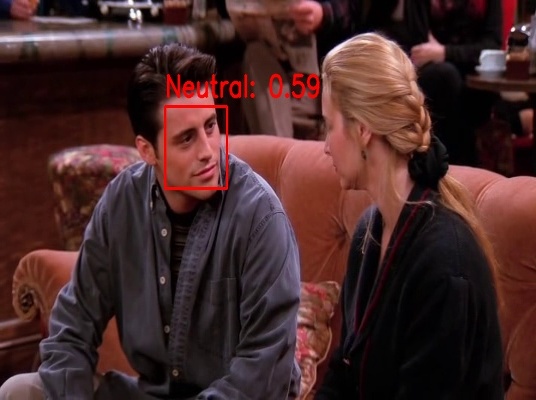}} &
\subfloat[(e) Neutral]{\includegraphics[width = 0.13\textwidth, height=0.14\textwidth]{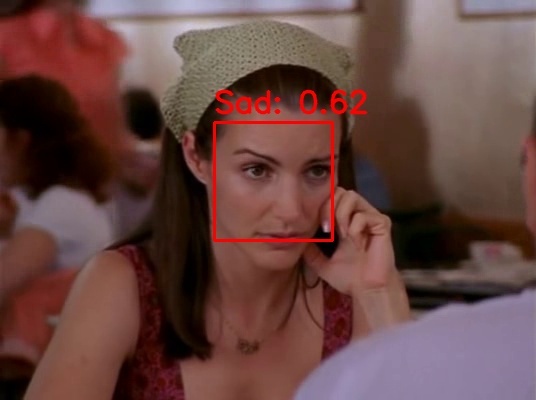}} &
\subfloat[(f) Sad]{\includegraphics[width = 0.13\textwidth, height=0.14\textwidth]{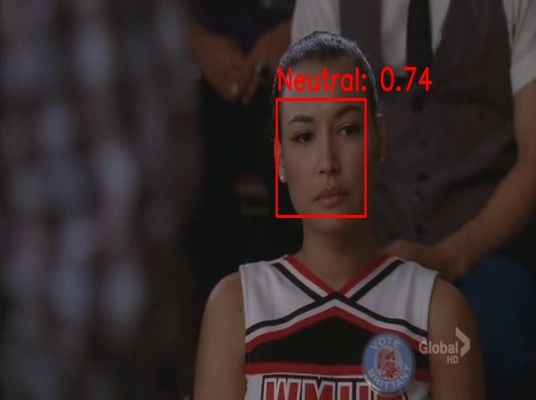}} &
\subfloat[(g) Surprise]{\includegraphics[width = 0.13\textwidth, height=0.14\textwidth]{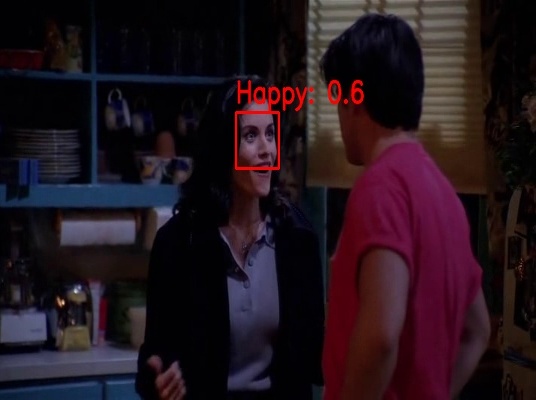}}
\\
\end{tabular}
\vspace{0ex}
\caption{Predictions on the NCAER-S test set. The first two rows (\textit{i}) and (\textit{ii}) show the results of the CAER-Net-S while the last two rows (\textit{iii}) and (\textit{iv}) demonstrate predictions of our GLAMOR-Net. The columns' names from (a) to (g) denote the ground-truth emotion of the images}
\label{fig:predictions}
\end{figure*}

\begin{figure*}
\centering
\setlength\tabcolsep{1.7pt} 
\begin{tabular}{ccc}
\subfloat{\includegraphics[width = 0.3\textwidth, height=0.22\textwidth]{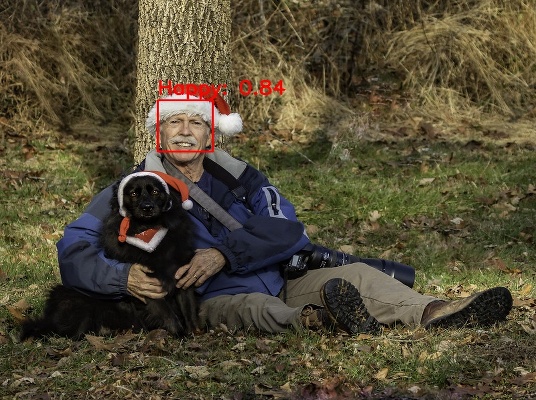}} &
\subfloat{\includegraphics[width = 0.3\textwidth,  height=0.22\textwidth]{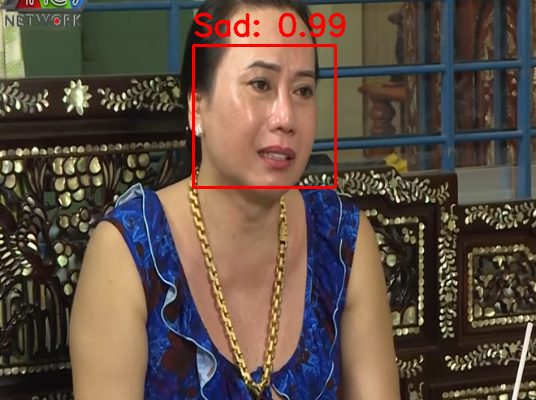}} & \subfloat{\includegraphics[width = 0.3\textwidth, height=0.22\textwidth]{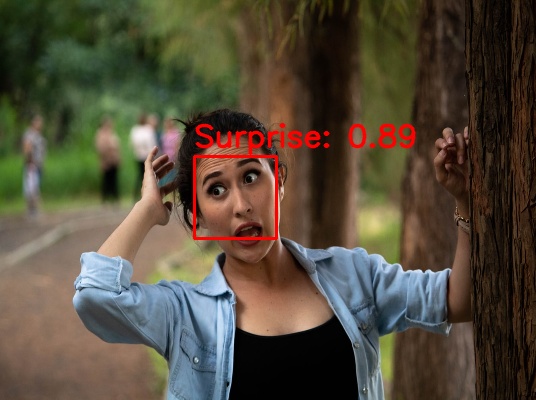}} \\
\subfloat{\includegraphics[width = 0.3\textwidth,  height=0.22\textwidth]{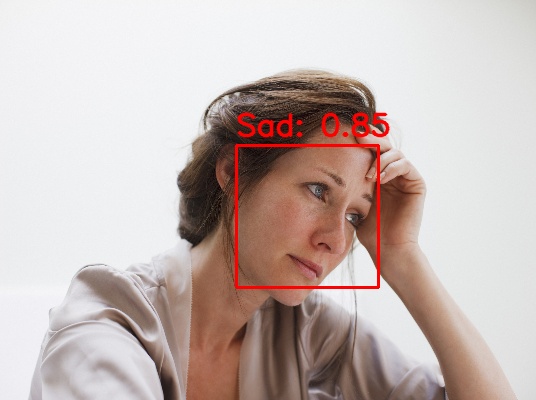}} &
\subfloat{\includegraphics[width = 0.3\textwidth,  height=0.22\textwidth]{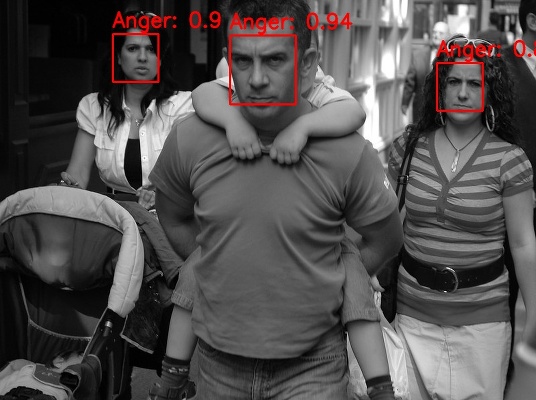}} &
\subfloat{\includegraphics[width = 0.3\textwidth,  height=0.22\textwidth]{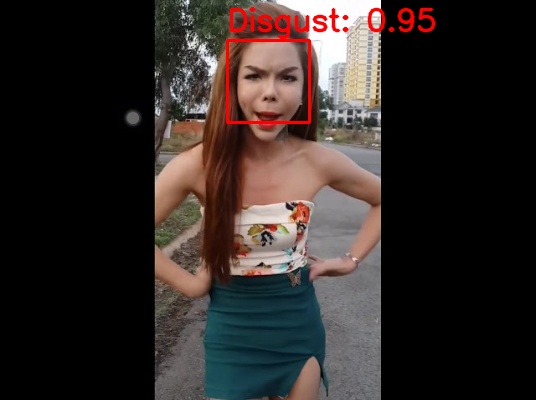}}
\end{tabular}
\vspace{0ex}
\caption{Human emotion detection results in the wild setting}
\label{fig:in_the_wild}
\end{figure*}

\subsubsection{Visualization}
Fig. \ref{fig:attention_map} shows the qualitative visualization with learned attention maps obtained by our method GLAMOR-Net in comparison with CAER-Net-S. It can be seen that our Global-Local attention mechanism produces better saliency maps and helps the model attend to the right discriminative regions in the surrounding background than the attention map produced by CAER-Net-S \cite{caer}. As we can see, our model is able to focus on the gesture of the person (Fig. \ref{fig:attention_map}f) and also the face of surrounding people (Fig. \ref{fig:attention_map}c, Fig. \ref{fig:attention_map}d) to infer the emotion accurately. 

Fig. \ref{fig:predictions} shows some emotion recognition results of different approaches on the NCAER-S dataset. More specifically, the first two rows (\textit{i}) and (\textit{ii}) contain predictions of the CAER-Net-S while the last two rows (\textit{iii}) and (\textit{iv}) show the results of our GLAMOR-Net.  In some cases, our model was able to exploit the context effectively to perform inference accurately. For instance, with the same \texttt{sad} image input (shown on the (\textit{i}) and (\textit{iii}) rows), the CAER-Net-S misclassified it as \texttt{neutral} while the GLAMOR-Net correctly recognized the true emotion category. It might be because our model was able to identify that the man was hugging and appeasing the woman and inferred that they were sad. Another example is shown on the (\textit{i}) and (\textit{iii}) rows of the \texttt{fear} column. Our model classified the input accurately, while the CAER-Net-S might be confused between the facial expression and the wedding surrounding, thus incorrectly predicted the emotion as \texttt{happy}.

On the other hand, we can also see on the (\textit{iv}) rows of Fig. \ref{fig:predictions}, the GLAMOR-Net misclassified the \texttt{disgust} and the \texttt{surprise} images as \texttt{happy} and the \texttt{neutral} image as \texttt{sad}. The reason might be that these images look quite confusing even to humans. Our model also failed to recognize emotions in the \texttt{anger}, \texttt{fear}, \texttt{happy} and \texttt{sad} images on the (\textit{iv}) rows and predicted them as \texttt{neutral} instead. It can be because the facial expression in these images does not manifest clearly enough, which makes it difficult to distinguish between the \texttt{neutral} class and these emotion categories. This uncertainty was previously shown in the confusion matrix in Fig. \ref{fig:ncaers_confusion_matrix}. 

\subsubsection{Emotion Recognition in The Wild}
As both the CAER-S dataset and its new split NCAER-S dataset contain only images from movie settings, they have a very limited number of people in a constrained environment. Therefore, the model trained using these datasets potentially do not work well on real-world image setting. Despite this challenge, Fig. \ref{fig:in_the_wild} shows that our GLAMOR-Net can successfully detect and recognize human emotion in these challenging settings. Note that, the input images in this setup do not share any overlap with the movie settings as in the training set. This again confirms the generalization ability of our proposed method.

%% file: 5_conclusion.tex
\section{Conclusions and Future Work}
\label{sec_conclusion}
In this work, we presented a novel method to exploit context information more efficiently by using the proposed global-local attention model. We have shown that our approach can considerably improve the emotion classification accuracy compared to the current state-of-the-art result in the context-aware emotion recognition task. The results on the CAER-S and the NCAER-S dataset consistently demonstrate the effectiveness and robustness of our method. 

Our approach currently only takes static images as input, which limits the amount of knowledge that can be exploited. We are planning to utilize temporal information in dynamic videos and other modalities such as audio in order to further improve the performance. We also consider releasing a more challenging emotion recognition dataset that contains rich background contexts with multiple faces in the same frame and take advantage of our attention model to extract the context saliency map for each face in a more effective manner. We hope that our work will pave the way for future work in which predicting the emotions of different people simultaneously is tackled.